\pdfoutput=1

\documentclass[11pt]{article}

\usepackage[final]{ACL2023}
\usepackage[ruled,linesnumbered]{algorithm2e}
\usepackage{algpseudocode}
\usepackage{times}
\usepackage{latexsym}
\usepackage{booktabs} 
\usepackage{amsmath}  
\usepackage{xcolor}
\usepackage[T1]{fontenc}
\usepackage{booktabs}    
\usepackage{xcolor}      
\usepackage{multirow}    
\usepackage{array} 
\usepackage[utf8]{inputenc}
\usepackage{graphicx}
\usepackage{microtype}
\usepackage{enumitem}
\usepackage{inconsolata}
\usepackage{pifont}
\usepackage{colortbl}

\usepackage{amssymb}
\graphicspath{{img/}}   

\title{When TableQA Meets Noise: A Dual Denoising Framework for\\ Complex Questions and Large-scale Tables}

\author{
  \textbf{Shenghao Ye\textnormal{\textsuperscript{1}\footnotemark[1]}}\ ,
  \textbf{Yu Guo\textnormal{\textsuperscript{1}\footnotemark[1]}}\ ,
  \textbf{Dong Jin\textnormal{\textsuperscript{3}\footnotemark[2]}}\ ,
  \textbf{Yuxiang Wang\textnormal{\textsuperscript{2}}}\ ,
  \textbf{Yikai Shen\textnormal{\textsuperscript{1}}},
  \textbf{Yunpeng Hou\textnormal{\textsuperscript{3}}},
  \textbf{Shuangwu Chen\textnormal{\textsuperscript{1}\footnotemark[2]}} \\
  \textbf{Jian Yang\textnormal{\textsuperscript{1}}},
  \textbf{Xiaofeng Jiang\textnormal{\textsuperscript{1}}}
\\
  \textsuperscript{1}University of Science and Technology of China \textsuperscript{2} The University of Melbourne
\\
  \textsuperscript{3}Institute of Artificial Intelligence, Hefei Comprehensive National Science Center
\\
  \texttt{\{ssh0321y, yukariguo, shenyikai, hyp314\}@mail.ustc.edu.cn}
\\
  \texttt{\{kingdon, chensw, jianyang, jxf\}@ustc.edu.cn}
}

\begin{document}
\maketitle
\renewcommand{\thefootnote}{\fnsymbol{footnote}}
\footnotetext[1]{Equal contribution}
\footnotetext[2]{Corresponding authors}
\begin{abstract}

Table question answering (TableQA) is a fundamental task in natural language processing (NLP). The strong reasoning capabilities of large language models (LLMs) have brought significant advances in this field. However, as real-world applications involve increasingly complex questions and larger tables, substantial noisy data is introduced, which severely degrades reasoning performance. To address this challenge, we focus on improving two core capabilities: Relevance Filtering, which identifies and retains information truly relevant to reasoning, and Table Pruning, which reduces table size while preserving essential content. Based on these principles, we propose EnoTab, a dual denoising framework for complex questions and large-scale tables. Specifically, we first perform Evidence-based Question Denoising by decomposing the question into minimal semantic units and filtering out those irrelevant to answer reasoning based on consistency and usability criteria. Then, we propose Evidence Tree-guided Table Denoising, which constructs an explicit and transparent table pruning path to remove irrelevant data step by step. At each pruning step, we observe the intermediate state of the table and apply a post-order node rollback mechanism to handle abnormal table states, ultimately producing a highly reliable sub-table for final answer reasoning. Finally, extensive experiments show that EnoTab achieves outstanding performance on TableQA tasks with complex questions and large-scale tables, confirming its effectiveness.

\end{abstract}

\begin{figure}[t]
    \centering
    \includegraphics[width=0.96\linewidth]{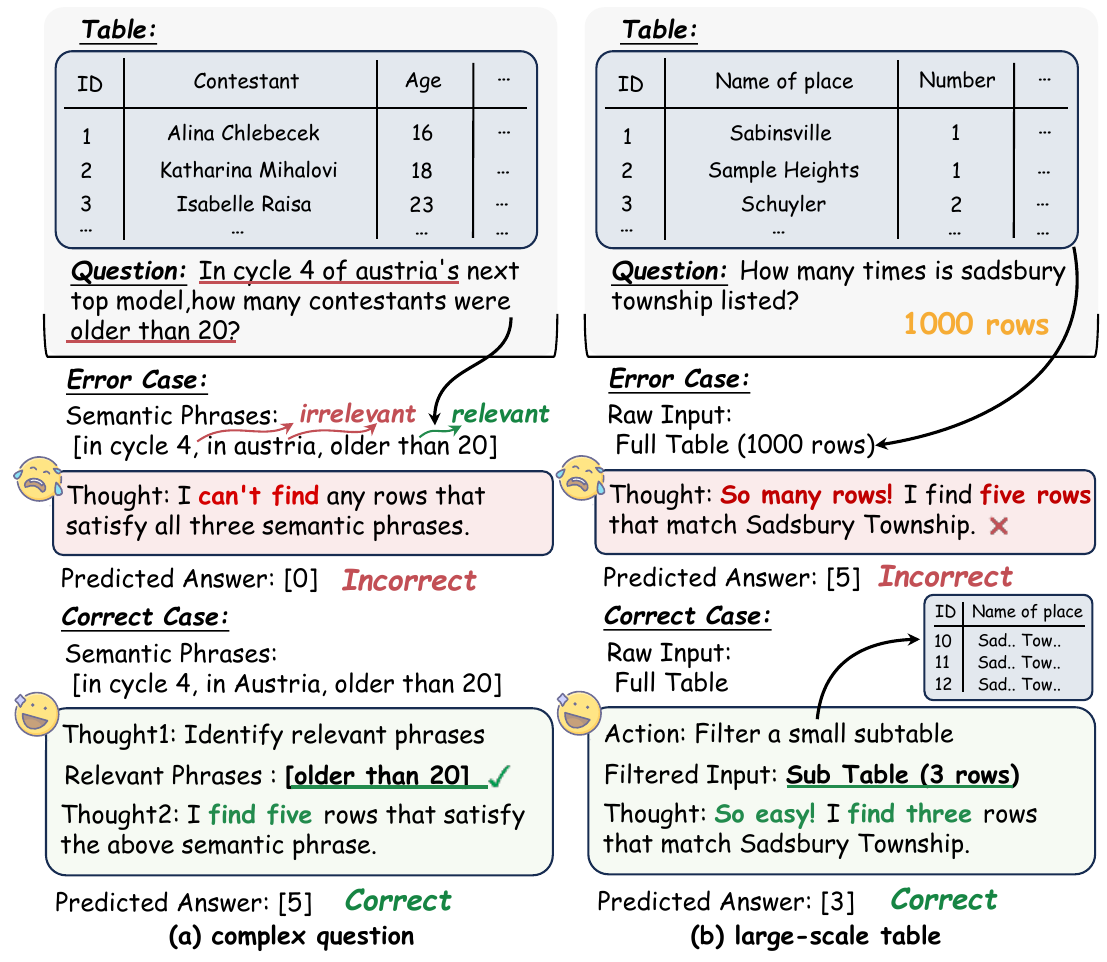}
    \caption{Error and Correct Cases for (a) the Complex Question and (b) the Large-Scale Table.}
    \label{fig:method1}
\end{figure}

\begin{figure*}[t]
    \centering
    \includegraphics[width=1\linewidth]{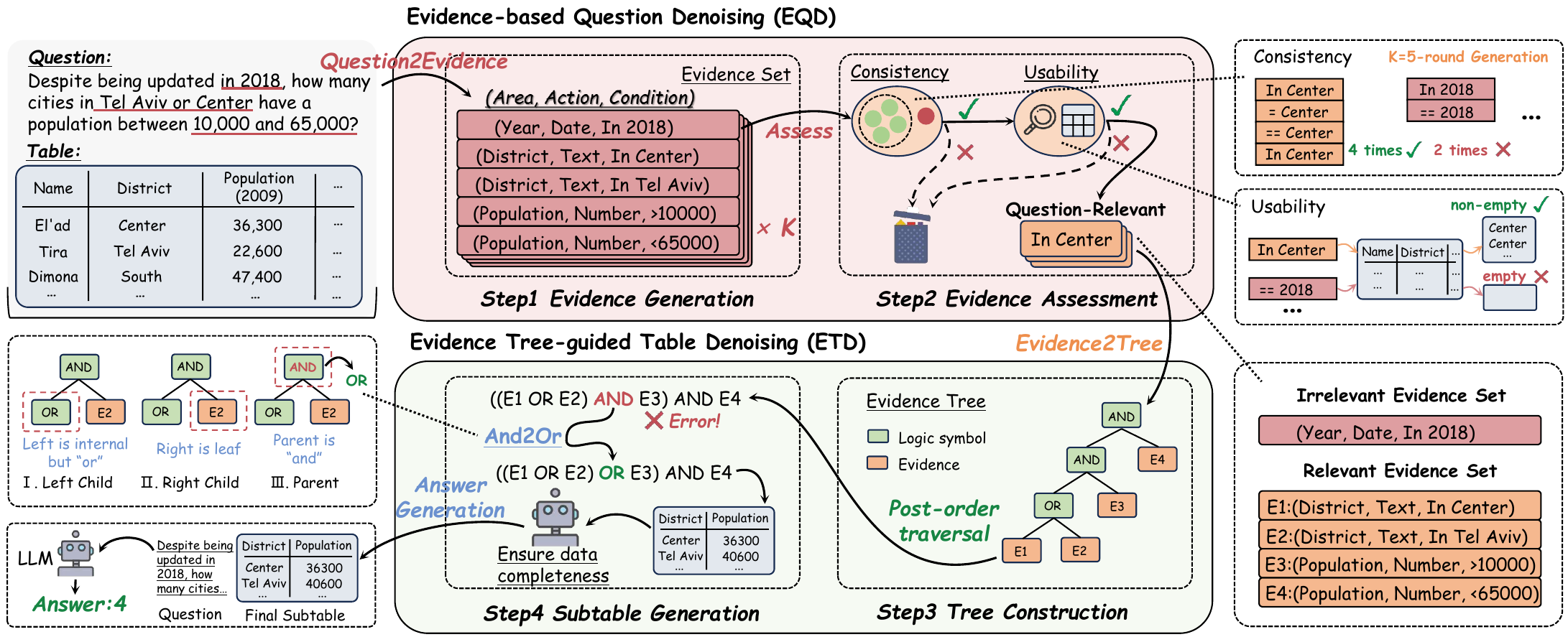}
    \caption{The EnoTab framework, composed of Evidence-based Question Denoising to remove irrelevant semantic units from the question and Evidence Tree-guided Table Denoising to eliminate irrelevant cell values from the table.}
    \label{fig:method2}
\end{figure*}

\section{Introduction}

Table question answering (TableQA) is a fundamental task in natural language processing (NLP) \citep{wang2022proton} that focuses on answering questions based on tabular data. With the rapid development of large language models (LLMs) \citep{openai2024gpt4technicalreport,grattafiori2024llama,team2024gemma}, this area has witnessed significant progress.  However, in real-world applications such as healthcare and finance, both the scale of tables and the complexity of questions are increasing \citep{cafarella2008webtables,zhang2025survey}, introducing substantial amounts of noisy data. Extensive research has demonstrated that such noisy data significantly degrades the performance of table reasoning \citep{chen2024tablerag,wang2025tabsd}.

\textit{Where noisy data Comes From and Why it Matters?} 
Noisy data in TableQA arises from both the question and the table \citep{zhou2025efficient,wang2025tabsd}. Questions often contain spurious correlations. As shown in Figure 1(a), phrases such as "in cycle 4" and "in Austria" are mistakenly treated by the model as constraints, though no corresponding data exists in the table, ultimately leading to an incorrect answer. Tables, meanwhile, often contain irrelevant data. As shown in Figure 1(b), only 3 of 260 rows contribute to answering the question, while the remaining 257 are entirely irrelevant, increasing reasoning difficulty and computational cost. In scenarios with complex questions and large-scale tables, noisy data becomes more pervasive and has greater impact on reasoning.

\textit{So, How to deal with noisy data?} We observe that effective TableQA under substantial noisy data requires two indispensable capabilities. (1) \textbf{Relevance Filtering}, i.e., identifying and ignoring spurious correlations in the question (e.g., "in cycle 4" and "in Austria"). As shown in Figure 1(a), once the model is explicitly informed which phrases are irrelevant and which relevant, it can easily derive the correct answer. (2) \textbf{Table Pruning}, i.e., removing irrelevant data to reduce the table size. As shown in Figure 1(b), when the original 260-row table is reduced to a 3-row subtable by retaining only the question-relevant data, the model can infer the correct answer more efficiently and accurately. These two capabilities effectively mitigate the impact of noisy data on table reasoning performance.

\textit{But why do existing methods fail?} Reviewing existing TableQA methods, we  summarize their limitations  into the following two aspects. First, it is hard to accurately identify spurious correlations in complex questions.
Some existing methods prompt the LLM to directly eliminate spurious correlations in complex questions in order to derive simpler subquestions \citep{ye2023large,wang2024chain,zhao2024tapera}, but this process is often error-prone, as the LLM tends to be overconfident and may mistakenly treat irrelevant elements as relevant. Other promising methods represent the question as a program and detect spurious correlations by checking whether the program executes correctly \citep{nahid2024tabsqlify,wang2025tabsd}. However, these correlations are often entangled with useful information, making it difficult to precisely locate and remove them even when their presence is detected. Secondly, it is easy to mistakenly lose target data relevant to answer reasoning when pruning the table. Most methods prune the table by generating SQL or Python programs \citep{zhang2023reactable,abhyankar2025h}. However, these methods often operate as black boxes, lacking transparency and making it hard to detect errors in time. Even if the loss of target data is identified, the pruning often has to be redone entirely \citep{wang2025tabsd}, resulting in increased overhead. In summary, two key challenges remain to be addressed: (1) how to accurately distinguish spurious correlations from truly relevant information in complex questions; and (2) how to effectively prune tables while preserving critical answer-related data.

To overcome these challenges, we propose EnoTab, which consists of two components: (1) \textbf{Evidence-based Question Denoising}, which introduces the concept of \textit{Evidence} to decompose a question into multiple minimal semantic units. Each unit is assessed independently based on two criteria: whether it appears consistently across multi-turn generation and whether corresponding data can be found in the table, in order to determine its relevance to answer reasoning. (2) \textbf{Evidence Tree-guided Table Denoising}, which constructs an \textit{Evidence Tree}, an explicit and transparent reasoning path for progressively pruning the table. Each pruning step is observable, allowing for timely detection of abnormal table states. To handle such abnormalities, we introduce a post-order node rollback mechanism to prevent the loss of answer-related data, ensuring that the final subtable remains reliable for supporting answer generation.

\textbf{Our Contributions}. (1) \textit{New Perspective}. We reveal and analyze the performance bottleneck of TableQA when handling complex questions and large-scale tables with substantial noisy data, and attribute it primarily to insufficient capabilities in \textbf{Relevance Filtering} and \textbf{Table Pruning}. (2) \textit{Novel Framework}. We propose EnoTab, a dual denoising framework designed to enhance noise robustness in TableQA tasks involving complex questions and large-scale tables. (3) \textit{SOTA Performance}. Extensive experimental results demonstrate the effectiveness of EnoTab, which achieves outstanding performance in TableQA tasks involving complex questions and large-scale tables.

\section{EnoTab}
\subsection{Problem Definition}
TableQA consists of a question $Q$, a table $T$, and the corresponding answer $Y$. The question $Q$ (e.g., "update in 2018, how many cities in Tel Aviv...?") can be represented as a set of semantic units $s_i$ (e.g., "in 2018", "cities in Tel Aviv"), with $Q = \{s_1, s_2, ..., s_n\}$. The table $T$ is represented as a two-dimensional matrix $T = \{c_{i,j} \mid 0 \le i < N,\ 0 \le j < M\}$, where $N$ is the number of rows, $M$ is the number of columns, and $c_{i,j}$ denotes the cell value at row $i$ and column $j$. The answer $Y=(y_1,y_2, ...,y_n)$. The goal is to infer the correct answer $Y$ given the question $Q$ and the table $T$:
\begin{equation}
Y = \arg\max \prod_{i=1}^{n} P_\theta(y_i \mid y_{<i}, Q, T; \theta)
\end{equation}
where $\theta$ denotes the parameters of a neural text generation model, and $y_i$ denotes the $i$-th tokens in the generated answer.

Note that not all semantic units $s_i$ and cell values $c_{i,j}$ contribute to deriving the answer $Y$. In this paper, we aim to explicitly identify and filter out such irrelevant data prior to reasoning, thereby producing a focused subset $E = \{s_1, s_3, \dots, s_n\} \subseteq Q$ and a subtable $T_\text{sub} \subseteq T$, thereby enabling more efficient and accurate answer derivation. 
\begin{equation}
Y = \arg\max \prod_{i=1}^{n} P_\theta(y_i \mid y_{<i}, E, T_\text{sub}; \theta)
\end{equation}

\subsection{Model Overview}
We propose EnoTab, a dual denoising framework designed to enhance noise robustness in TableQA tasks involving complex questions and large-scale tables. As shown in Figure~\ref{fig:method2}, EnoTab is composed of the Evidence-based Question Denoising (EQD) and the Evidence Tree-guided Table Denoising (ETD). EQD is responsible for removing irrelevant semantic units from the question, while ETD focuses on eliminating irrelevant cell values from the table. More specifically, EQD decomposes the question into a set of evidences, each representing a minimal semantic unit. It then assesses their relevance to answer reasoning based on two criteria, producing a relevant evidence set $E_r$. Based on this set, ETD builds an explicit table pruning path, called the Evidence Tree. Each step is observable, allowing real-time monitoring and enabling a post-order rollback mechanism to prevent the loss of answer-related data, ultimately yielding a reliable subtable $T_{\text{sub}}$. Finally, EnoTab generates the answer $Y$ to question $Q$ based on the evidence set $E_r$ and the subtable $T_{\text{sub}}$. We next provide a detailed introduction to EQD (Sec.~\ref{sec:stage1}) and ETD (Sec.~\ref{sec:stage2}), 
with further implementation details and hyperparameters available in Appendix~\ref{app:impl}.

\subsection{Evidence-based Question Denoising}
\label{sec:stage1}
Complex questions typically contain more semantic units, and both relevant and irrelevant units often appear semantically related to the question, making them hard to distinguish. Therefore, directly identifying irrelevant units based on the entire question is often inaccurate \citep{ye2023large}. To address this, we design \textit{Evidence}, defined as follows.
\paragraph{Definition 1}
\textbf{\textit{Evidence}} is defined as the representation of a minimal semantic unit in the question that is aligned with a specific data region in the table. Each evidence is formulated as a triplet $e = (\textit{area}, \textit{condition}, \textit{action})$, where $\textit{area}$ specifies the column in the table associated with the semantic unit, $\textit{condition}$ denotes the cell values in that column satisfying the semantic unit, and $\textit{action}$ indicates how the condition should be applied (e.g., string matching, numerical comparison, or date evaluation). For example, for the unit "cities in Tel Aviv", the corresponding evidence links to the \texttt{District} column, sets the condition to \texttt{Tel Aviv}, and uses \texttt{string matching} as the action.

\paragraph{Evidence Generation}
Unlike subquestions or program code that often involve multiple semantic units \citep{ye2023large,nahid2024tabsqlify}, each evidence corresponds to only one unit, which means it is simpler in structure and easier to generate. Specifically, for a complex question $Q$, we leverage a powerful LLM $M_e$ to generate an evidence set $E$ by decomposing $Q$, formulated as:
\begin{equation}
E = \{e_1, e_2, \dots, e_n\} \gets M_e(Q, H, R)
\end{equation}
where $H$ represents the table header, $R$ denotes $k$ representative rows sampled from the table $T$. These rows serve as context to help $M_e$ understand the table schema and semantics during evidence generation. Inspired by Chase-SQL \citep{pourreza2024chase}, we adopt a two-stage retrieval process. First, an LLM extracts keywords from the question to perform coarse filtering via Locality-Sensitive Hashing (LSH). The retrieved candidates are then re-ranked using embedding similarity and edit distance to select the top-$k$ most relevant rows.

Note that traditional methods focus on using question decomposition to plan reasoning paths \citep{ye2023large,wang2024chain}, while we aim to extract individual semantic units from the question with decomposition as a principled technique, in order to facilitate subsequent relevance assessment.

\paragraph{Evidence Evaluation}
Given the evidence set $E$, a natural problem arises: how to assess whether each evidence is relevant to answer reasoning. However, unlike simple questions with ground truth, evidence requires more reliable assessment criteria, which means that approaches such as only using the LLM as a judge are no longer suitable \citep{lin2025explore}. Therefore, we design a more comprehensive assessment criterion based on two aspects: consistency and usability.

Specifically, inspired by self-consistency \citep{wang2022self}, we assume that evidence should appear consistent during the multi-round generation if it is truly relevant to answer reasoning. So, how can we determine whether there is consistent evidence across multi-round generations? Each evidence is defined as a triple $(\textit{area}, \textit{condition}, \textit{action})$ (see Definition 1), where $\textit{area}$ and $\textit{action}$ can be directly compared via string match. However, the same \textit{condition} may appear in different surface forms. For example, in Figure 2, "in Center" and "== Center" are considered consistent. To address this, we use a semantic discriminator $M_d$ such as LLaMA \citep{grattafiori2024llama} to determine whether two conditions are semantically equivalent:
\begin{equation}
\textit{same} \gets M_d(\textit{condition}_1, \textit{condition}_2)
\end{equation}
where $\textit{same} \in \{\text{True}, \text{False}\}$. Then we generate $n$ rounds of evidence sets and design an efficient algorithm (see Algorithm \ref{alg:evidence_consistency} in the Appendix \ref{appendix:A.2}) to generate a candidate evidence set and compute the consistency score $S$ for each candidate. If $S \ge \alpha$, where $\alpha$ is a predefined threshold, the evidence is retained; otherwise, it is discarded.

How do we define the usability of evidence? As stated in Definition 1, each evidence is expected to be grounded in the table. We consider an evidence instance unusable for answer reasoning if it cannot be matched with any corresponding data in the table, regardless of whether the failure arises from semantic irrelevance or system limitations. To support this process, we design a toolkit $\mathcal{P}$ that integrates multiple APIs to determine whether a given piece of evidence can be grounded in the table. Given an evidence $e$ and a table $T$ as input, $\mathcal{P}$ searches for data in $T$ that satisfies $e$. If a match is found, it returns \texttt{true}; otherwise, it returns \texttt{false}. Only evidences for which $\mathcal{P}$ returns \texttt{true} are retained for subsequent reasoning. Based on the two criteria described above, we obtain a reliable evidence set $E_r$.

\subsection{Evidence Tree-guided Table Denoising}
\label{sec:stage2}
To further improve the efficiency and effectiveness of reasoning, it is also necessary to remove data in the table that is irrelevant to answer reasoning. Most existing table pruning methods operate in a black-box manner, which increases the risk of losing target data. Therefore, we design the \textit{Evidence Tree}, which constructs an explicit and transparent pruning path based on the reliable evidence set $E_r$ as the filtering criterion. The \textit{Evidence Tree} is defined as follows:

\paragraph{Definition 2}  \textbf{\textit{Evidence Tree}} is a binary tree (as illustrated in Step 3 of Figure \ref{fig:method2}), denoted as $\mathcal{T} = (N_{\text{leaf}},N_{\text{inter}})$. $N_{\text{leaf}}$ is the set of leaf nodes, where each node satisfies $n_{\text{leaf}} \in E_r$, meaning that each leaf corresponds to one element in the evidence set. Each leaf node takes the original table as input, filters it based on the corresponding evidence, and outputs a sub-table. $N_{\text{inter}}$ is the set of internal nodes, where each node satisfies $n_{\text{inter}} \in \{\text{And}, \text{Or}\}$, indicating the logical relation between its two child nodes. Each internal node takes the two sub-tables from its child nodes as input and outputs a merged sub-table according to the specified logical relation.

\paragraph{Tree Construction}
We also employ a powerful LLM $M_r$ to generate the evidence tree. Given the table header $H$, the same $k$ representative rows $R$, the question $Q$, and a reliable evidence set $E_r$, we prompt $M_r$ to produce the evidence tree $\mathcal{T}$. Formally,
\begin{equation}
\mathcal{T} = (N_{\text{leaf}},N_{\text{inter}}) \gets M_r(Q, H, R, E)
\end{equation}
Further implementation details are provided in the Appendix \ref{app:impl}. Since the evidence tree $\mathcal{T}$ is a binary tree, we can adopt post-order traversal to linearize the execution of this inherently non-linear structure. To execute this process, we employ the previously introduced toolkit \( \mathcal{P} \). For a leaf node \( n_{\text{leaf}} \), \( \mathcal{P} \) takes the original table \( T \) and the associated evidence \( e_r \) as input, and outputs a corresponding sub-table. For an internal node \( n_{\text{inter}} \), \( \mathcal{P} \) takes the two sub-tables generated by its left and right child nodes, along with the logical operator contained in \( n_{\text{inter}} \), and returns a merged table based on that operator. Finally, our goal is to obtain a sub-table \( T_{\text{sub}} \) by feeding the original table \( T \) and the evidence tree $\mathcal{T}$ into the toolkit \( \mathcal{P} \).

\paragraph{Subtable Generation}

During the post-order traversal of the evidence tree \( \mathcal{T} \), the resulting sub-table \( T_{\text{sub}} \) may sometimes be empty. To address this issue, we first analyze the source of the problem. It typically arises at internal nodes \( n_{\text{inter}} \) with the \texttt{AND} operator, since each leaf node \( n_{\text{leaf}} \) is guaranteed to produce non-empty tables through evidence assessment, and internal nodes \( n_{\text{inter}} \) with the \texttt{OR} operator do not eliminate all entries.
We then further examine its underlying causes: (1) critical answer-related data has been lost prior to reaching the current node \( n_{\text{inter}} \); or (2) the \texttt{AND} operation at the current node \( n_{\text{inter}} \) yields an empty intersection. To mitigate this, we propose a simple yet effective method called the \textit{And2Or} operation, which replaces \texttt{AND} with \texttt{OR}. Specifically, we apply the \textit{And2Or} operation sequentially to the left child, right child, and finally the current node. The procedure (illustrated in Figure~\ref{fig:method2}) proceeds as follows: (i) check whether the current node is an internal node with \texttt{AND} logic; (ii) if yes, replace \texttt{AND} with \texttt{OR}; (iii) if a non-empty table is produced, stop; if not, move to the next node. Finally, we ensure that a non-empty subtable $T_\text{sub}$ is obtained.

To further guarantee the completeness of the target data, we employ a table verifier $M_i$ to determine whether $T_{\text{sub}}$ contains all the information required to answer $Q$. Given $T_{\text{sub}}$ as input, $M_i$ returns \texttt{True} or \texttt{False}. If the result is \texttt{True}, $T_{\text{sub}}$ is accepted as the final subtable. Otherwise, the process rolls back to the subtable generated at the previous node and re-initiates verification. For cost efficiency, we allow at most two verification attempts. If the second attempt still returns \texttt{False}, the full table is used instead. Notably, traditional methods typically regenerate the subtable from scratch when encountering incomplete information \citep{yu-etal-2025-table}, while our approach enhances efficiency by reusing the previous subtable rather than restarting the process. At the end, we obtain the final subtable $T_{\text{final}}$.

\paragraph{Answer Generation} We have removed irrelevant content in the table $T$ and the question $Q$. In this stage, we perform end-to-end TableQA. Given a subtable $T_{\text{final}}$ and a highlighted question $Q_{\text{final}}$ using the relevant evidence set $E_r$, we prompt the LLM to generate the final answer $Y$.

\begin{table*}[htbp]
\centering
\resizebox{\textwidth}{!}{
\begin{tabular}{l|cc|cc|c}
\toprule[1pt]
\multirow{2}{*}{\textbf{Method}} 
& \multicolumn{2}{c|}{\textbf{STQA-N}} 
& \multicolumn{2}{c|}{\textbf{STQA-L}}
& \multirow{2}{*}{\textbf{Average}} \\
\cmidrule(lr){2-3} \cmidrule(lr){4-5}
& \textbf{GPT-4o} & \textbf{GPT-4o-mini}
& \textbf{GPT-4o} & \textbf{GPT-4o-mini} \\
\midrule
End-to-End QA  &  62.8 & 57.6 &  66.4 & 59.8 & 61.7 \\
Chain-of-Thought & 63.7 & 59.8 & 66.7 & 60.5 & 62.7\\
Text-to-SQL \citep{rajkumar2022evaluating}  & 61.7 & 62.6 & 46.3 & 47.1 & 54.4\\
Binder \citep{Binder} & 64.1 & 63.9 & 48.7 & 45.6 & 55.6\\
Dater \citep{ye2023large}& 59.3 & 54.4 & 50.4 & 42.6 & 51.7 \\
Chain-of-Table \citep{wang2024chain} & 59.7 & 56.9 & 57.1 & 57.3 & 57.8 \\
TabSQLify \citep{nahid2024tabsqlify} & 65.9 & 61.8 & 60.6 & 58.7 & 61.8 \\
H-Star \citep{abhyankar2025h}& 70.2 & 68.6 & 66.2 & 63.8 & 67.2\\
TabLaP \citep{wang2025accurate} & \underline{73.6} & \underline{70.8} & \underline{69.1} & \underline{65.9} & \underline{69.9} \\
\midrule
\rowcolor[HTML]{D8ECE4}
\textbf{EnoTab(Ours)} &
\textbf{80.3 \textcolor[HTML]{076B3D}{(+8.3)}} & 
\textbf{78.2 \textcolor[HTML]{076B3D}{(+9.5)}} & 
\textbf{75.3 \textcolor[HTML]{076B3D}{(+8.2)}} & 
\textbf{72.5 \textcolor[HTML]{076B3D}{(+9.1)}} & 
\textbf{76.6 \textcolor[HTML]{076B3D}{(+8.7)}}\\
\bottomrule[1pt]
\end{tabular}%
}
\caption{Overall results on two large-scale TableQA datasets STQA-N and STQA-L. The best result is \textbf{bold}, and the second-best result is \underline{underlined}. Numbers in () indicate the performance gain over the second-best method.}
\label{tab:size_analysis}
\vspace{-14pt}
\end{table*}

\section{Experiments}

\subsection{Experimental Setup}
\paragraph{Datasets} We evaluate EnoTab on four datasets: WikiTQ \citep{pasupat2015compositional}, a table reasoning dataset with 4,344 samples from 421 tables, and TabFact \citep{2019TabFactA}, a table-based fact verification dataset containing 2,024 samples from 298 tables. In addition, we construct two large-scale table datasets based on Spider \citep{yu2018spider}, denoted as STQA-L and STQA-N, corresponding to naturally large tables and noise-injected large tables \citep{wang2025tabsd} (construction details are provided in the Appendix~\ref{apx:data}). 


\paragraph{Baselines}
We compare our approach against three categories of baselines:
(1) \textbf{Generic methods} End-to-End QA, Chain-of-Thought, and Text-to-SQL \citep{rajkumar2022evaluating}.
(2) \textbf{Decomposition-based methods} Dater \citep{ye2023large}, Chain-of-Table \citep{wang2024chain}, and TabLaP \citep{wang2025accurate}.
(3) \textbf{Pruning-based methods} Binder \citep{Binder}, TabSQLify \citep{nahid2024tabsqlify}, and H-Star \citep{abhyankar2025h}.


\paragraph{Evaluation Metric} For the STQA-N, STQA-L, and WikiTQ datasets, we use exact match accuracy to check whether the predicted answer matches the ground truth. For TabFact, we adopt binary classification accuracy as evaluation metric.


\subsection{Main Results}

\paragraph{Large-scale Tables}
Table \ref{tab:size_analysis} presents the performance of EnoTab on two large-scale table datasets: STQA-N and STQA-L. The results show that EnoTab significantly outperforms all baseline methods, demonstrating its effectiveness in large-scale TableQA scenarios. On STQA-N, both decomposition-based and pruning-based methods achieve relatively good performance, with pruning-based methods performing slightly better. We attribute this to the high volume of noisy tokens in STQA-N, which degrades reasoning performance. In contrast, pruning-based methods help alleviate reasoning pressure by removing irrelevant data. Among them, EnoTab stands out by leveraging its powerful and stable pruning capability, achieving state-of-the-art performance.

Furthermore, EnoTab exhibits even greater advantages on the STQA-L dataset, further demonstrating its superiority on truly large-scale tables. The results show that EnoTab achieves more significant performance gains, while pruning-based methods suffer a notable performance drop. We attribute this to the fact that STQA-L consists of naturally large tables with more complex structures and data formats compared to STQA-N, thus requiring more precise pruning. Poor pruning often leads to the loss of answer-related information, resulting in reasoning failures. In contrast, question decomposition methods maintain relatively stable performance, benefiting from the strong reasoning capabilities of the underlying language models. These findings further validate  EnoTab’s exceptional capability in handling large-scale tables.

\begin{figure}[t]
    \centering
    \includegraphics[width=1.0\linewidth]{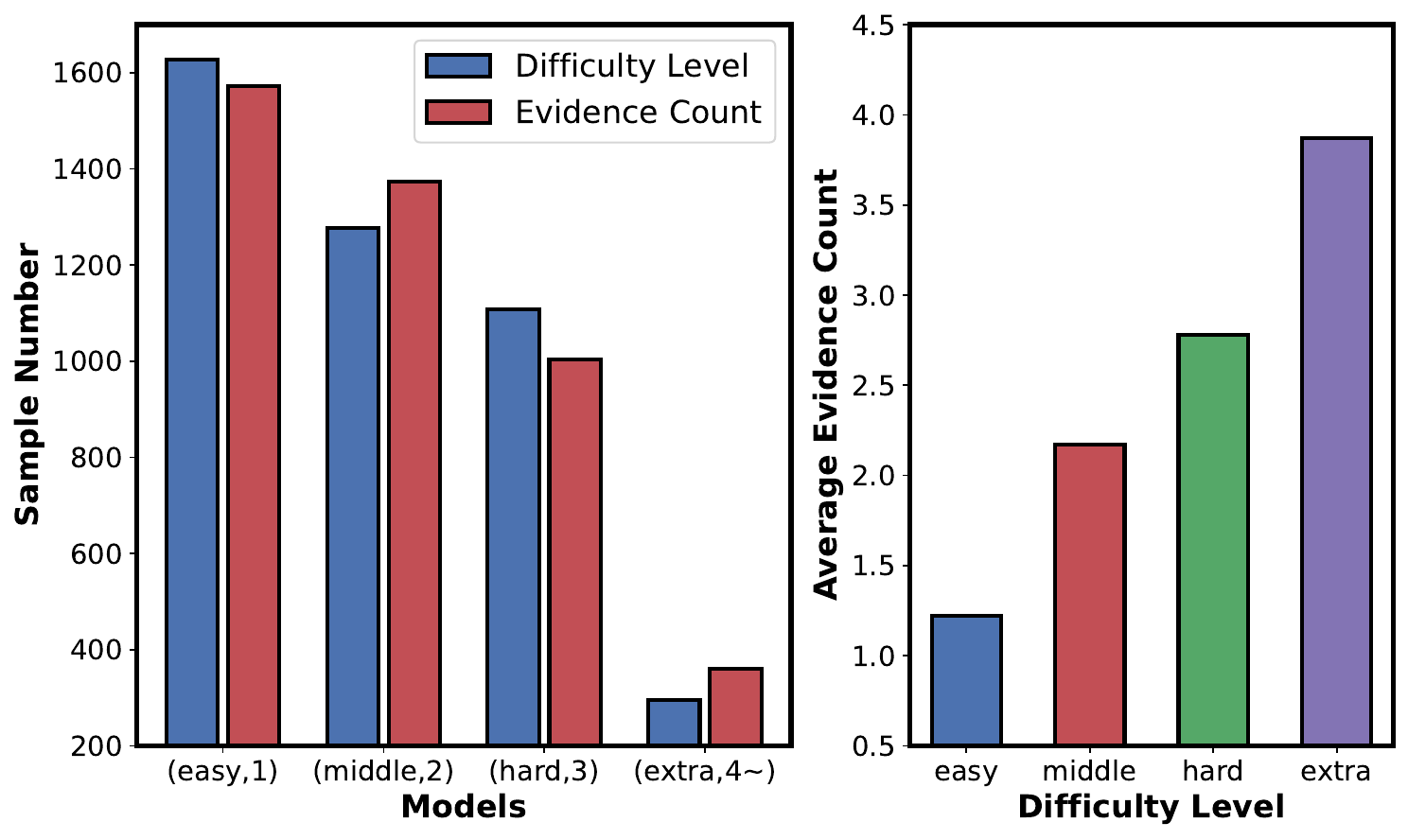}
    \caption{Distribution of question difficulty levels in the WikiTQ dataset and the average number of evidences generated by EnoTab per level.}
    \label{fig:chain_length2}
\end{figure}

\paragraph{Complex Questions} 
To evaluate EnoTab’s ability to handle complex questions, we categorize questions in the WikiTQ dataset into difficulty levels based on the reasoning performance of GPT-4o. Each question is independently answered 100 times by GPT-4o and labeled as “Easy” (90–100 correct), “Medium” (60–89), “Hard” (10–59), or “Extra Hard” (0–9). The distribution across difficulty levels is shown in Figure~\ref{fig:chain_length2}.
We further analyze the number of evidences generated by EnoTab for questions of different difficulty levels, along with the average per level. Results reveal a clear positive correlation: the more complex the question, the more evidences EnoTab generates, indicating that the method adaptively adjusts its reasoning depth according to question difficulty.

\begin{figure}[t]
    \centering
    \includegraphics[width=1.0\linewidth]{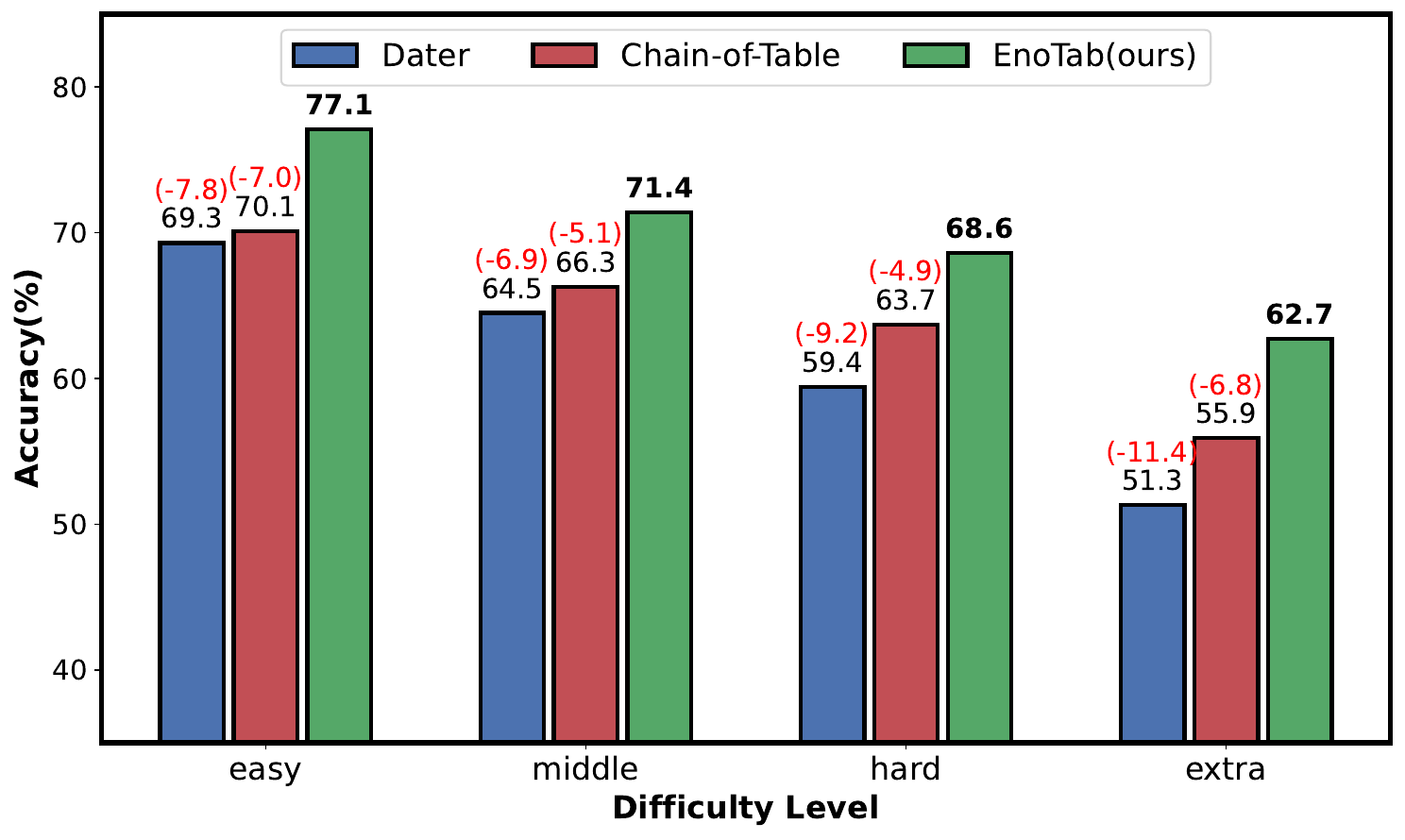}
    \caption{Accuracy comparison of EnoTab, Chain-of-Table, and Dater across different difficulty levels in the WikiTQ dataset, evaluated with GPT-4o-mini.}
    \label{fig:chain_length}
\end{figure}

Figure~\ref{fig:chain_length} shows the performance comparison between EnoTab and typical decomposition-based methods such as Chain-of-Table and Dater under different difficulty levels using GPT-4o-mini. EnoTab consistently achieves higher accuracy across all levels. In particular, under the “Extra Hard” setting, where baseline performance drops significantly, EnoTab maintains a clear advantage. These results further validate the robustness and reasoning ability of EnoTab in handling complex questions.

\paragraph {Standard Benchmark} 
We further evaluate EnoTab on two standard benchmark datasets, WikiTQ and TabFact, using GPT-4o-mini as the evaluation model. Table \ref{tab:main_exp} shows that EnoTab consistently outperforms existing methods on both datasets. This demonstrates that EnoTab not only excels in substantial noisy data scenarios such as complex questions and large-scale tables, but also maintains strong performance on standard TableQA tasks, indicating good generalizability.

\begin{table}[t]
\centering
\resizebox{0.48\textwidth}{!}{
\begin{tabular}{l|c|c|c}
\toprule[1pt]
\textbf{Method}
& \textbf{WikiTQ}
& \textbf{TabFact}
& \textbf{Average}\\
\midrule
End-to-End QA   & 52.6  & 73.5 & 63.1 \\
Chain-of-Thought  & 58.2  & 77.2 & 67.7\\
Text-to-SQL   & 52.9  & 69.7 &61.3\\
Binder  & 58.8  & 77.2 &68.0\\
Dater  & 58.3  & 80.1 &69.2 \\
Chain-of-Table   & 67.1 & 84.2 &75.7\\
TabSQLify  & 66.4  & 78.8 &72.6 \\
H-Star  & \underline{73.8} & \underline{88.4} &\underline{81.1}\\
TabLaP  & 72.8  & 86.9& 79.9 \\
\midrule
\rowcolor[HTML]{D8ECE4}
\textbf{EnoTab(Ours)} &
\textbf{74.6 \textcolor[HTML]{076B3D}{(+1.8)}} & 
\textbf{89.2 \textcolor[HTML]{076B3D}{(+2.3)}} &
\textbf{81.9 \textcolor[HTML]{076B3D}{(+2.0)}}\\
\bottomrule[1pt]
\end{tabular}%
}
\caption{Performance comparison between EnoTab and previous work on WikiTQ and TabFact datasets. The best result is \textbf{bold}, and the second-best result is \underline{underlined}.}
\label{tab:main_exp}
\end{table}

\begin{table}[t]
\centering
\resizebox{0.5\textwidth}{!}{
\begin{tabular}{l|cc|cc}
\toprule
\textbf{Method} & \textbf{STQA-N} & $\triangledown$ & \textbf{STQA-L} & $\triangledown$ \\
\midrule
\textbf{EnoTab(GPT-4o)} & \textbf{80.3} & -- & \textbf{75.3} & -- \\
\quad w/o Consistency Assessment & 74.0 & \textcolor[HTML]{CC0000}{(-6.3)} & 69.7 & \textcolor[HTML]{CC0000}{(-5.6)} \\
\quad w/o Usability Assessment & 72.1 & \textcolor[HTML]{CC0000}{(-8.2)} & 68.5 & \textcolor[HTML]{CC0000}{(-6.8)} \\
\quad w/o And2Or Operation & 76.1 & \textcolor[HTML]{CC0000}{(-4.2)} & 71.5 & \textcolor[HTML]{CC0000}{(-3.8)} \\
\quad w/o Table Verifier & 75.6 & \textcolor[HTML]{CC0000}{(-4.2)} & 72.8 & \textcolor[HTML]{CC0000}{(-2.5)} \\
\midrule
\textbf{EnoTab(GPT-4o-mini)} & \textbf{78.2} & -- & \textbf{72.5} & -- \\
\quad w/o Consistency Assessment    & 72.7 & \textcolor[HTML]{CC0000}{(-5.5)} & 66.1 & \textcolor[HTML]{CC0000}{(-6.4)} \\
\quad w/o Usability Assessment & 70.3 & \textcolor[HTML]{CC0000}{(-6.9)} & 65.6 & \textcolor[HTML]{CC0000}{(-6.9)} \\
\quad w/o And2Or Operation & 73.8 & \textcolor[HTML]{CC0000}{(-4.4)} & 68.1 & \textcolor[HTML]{CC0000}{(-4.4)} \\
\quad w/o Table Verifier  & 74.3 & \textcolor[HTML]{CC0000}{(-3.9)} & 68.5 & \textcolor[HTML]{CC0000}{(-4.0)} \\
\bottomrule
\end{tabular}
}
\caption{Ablation study. Evaluations are conducted on STQA-N and STQA-L using GPT-4o and GPT-4o-mini.}
\label{tab:ablation}
\end{table}

\subsection{Experimental Analysis}
\paragraph{Ablation Study}

To quantify the contribution of each component in EnoTab, we conduct ablation studies, as shown in Table~\ref{tab:ablation}. Specifically, we systematically disable the following key modules: \textit{Consistency Assessment} and \textit{Usability Assessment} from the Evidence Assessment module, and \textit{And2Or Operation} and \textit{Table Verifier} from the Subtable Generation module. Evaluations are performed using GPT-4o-mini on the STQA-L and STQA-N datasets.
The results indicate that disabling any single component leads to performance degradation on both datasets, confirming the significant positive impact of each module on the overall effectiveness of the framework. In particular, removing Consistency Assessment or Usability Assessment results in the most noticeable drop, highlighting their critical role in enhancing the framework’s relevance filtering capability by accurately identifying and eliminating spurious correlations in the question.
Likewise, removing the And2Or Operation or the Table Verifier also causes a substantial decline in performance, demonstrating their importance in improving the framework’s table pruning capability and in preventing the inadvertent removal of crucial information during pruning.
In summary, the results show that all components in EnoTab play indispensable roles in maintaining system performance. The absence of any module weakens the system’s reasoning ability, underscoring the synergy among components and the necessity of their integration for achieving robust performance.

\paragraph{Adaptability}

\begin{figure}[t]
    \centering
    \includegraphics[width=0.92\linewidth]{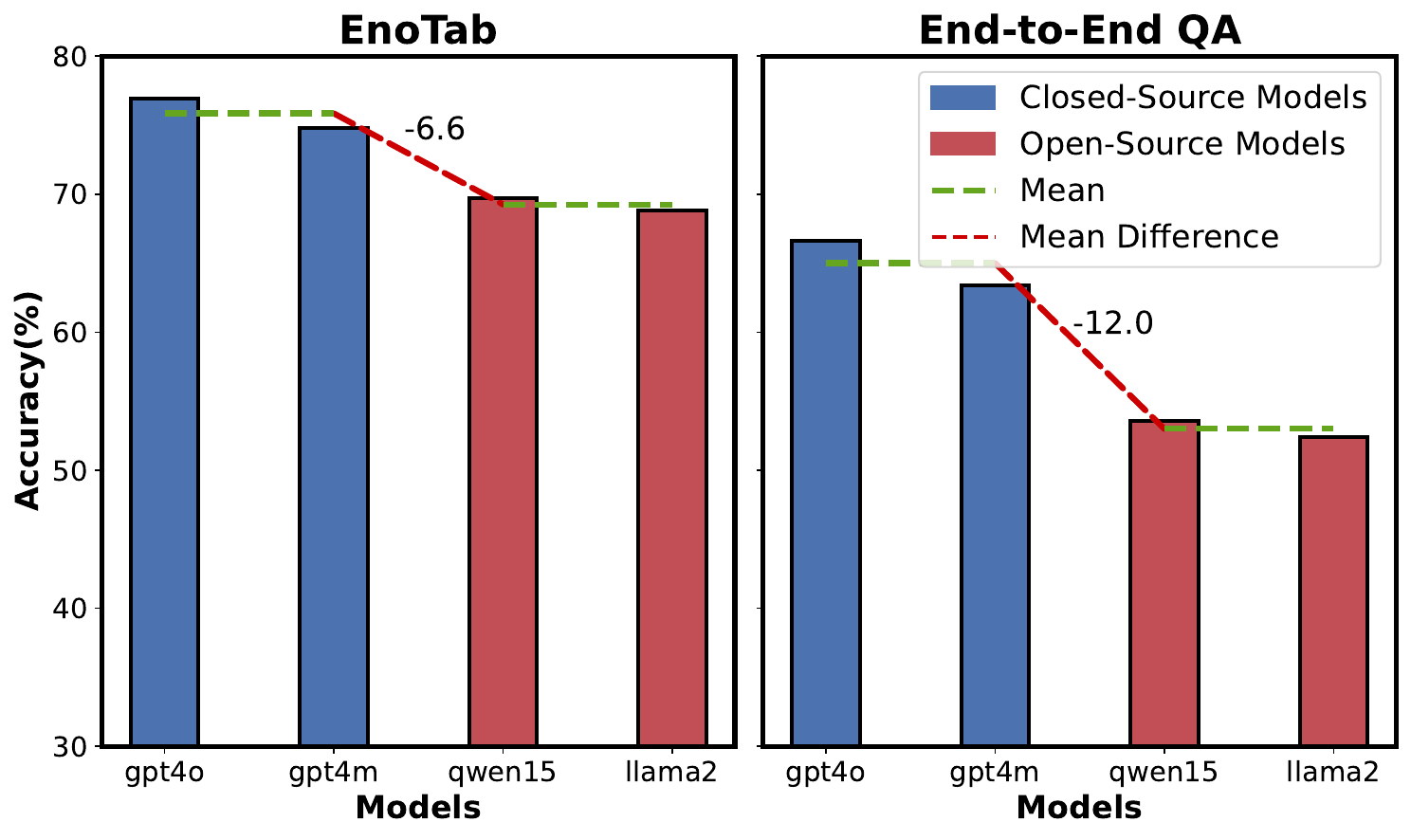}
    \caption{Execution accuracy of closed-source and open-source models on WikiTQ.}
    \label{fig:adaptability}
\end{figure}

To evaluate the adaptability of EnoTab across different types of foundation models, we conduct experiments using both closed-source models (GPT-4o-mini and GPT-4o) and open-source models (LLaMA-2-70B and Qwen-1.5-70B). We randomly sample 200 examples from the WikiTQ dataset and compare the performance of EnoTab with a standard End-to-End QA method. As shown in Figure~\ref{fig:adaptability}, End-to-End QA suffers a noticeable performance drop when switching from closed-source to open-source models, indicating a strong reliance on the underlying model's reasoning capability. In contrast, EnoTab maintains stable performance across different model configurations, with only minimal degradation even when paired with weaker models.
This robustness can be attributed to EnoTab’s effective relevance filtering and reliable table pruning capabilities, which jointly eliminate irrelevant content from both the table and the question. By reducing reasoning burden and complexity, EnoTab enables the model to perform well even under limited capacity. These findings suggest that EnoTab generalizes better across foundation models of varying quality and is more suitable for real-world scenarios where high-performance closed-source models may not always be available.

\paragraph{Table Pruning Effectiveness}
To evaluate the effectiveness of EnoTab's table pruning capability, we randomly sample 200 correctly answered instances from each of the four datasets and compute the average number of tokens before and after pruning. Table~\ref{tab:compression} reports the average token counts and corresponding compression rates. For the large-scale datasets STQA-L and STQA-N, EnoTab achieves substantial compression, demonstrating strong pruning ability by effectively removing irrelevant content. Meanwhile, on the relatively smaller-scale datasets WikiTQ and TabFact, EnoTab maintains reliable performance, avoiding the erroneous removal of answer-related data despite the limited presence of noise. These results confirm that EnoTab consistently adapts to varying table scales and exhibits robust and effective table pruning across different scenarios.

\section{Related Work}
The strong reasoning capabilities of LLMs have significantly advanced the development of TableQA. To further improve table reasoning performance, existing research has primarily explored two key directions (due to space limitations, more related work is discussed in the Appendix~\ref{apx:related_work}).
\paragraph{Question Decomposition}
This line of research typically decomposes complex questions into simpler intermediate steps to extend the reasoning chain, thereby easing the burden on LLMs and improving accuracy  \citep{ye2023large,wang2024chain,zhao2024tapera,wu2024protrix}. However, these methods heavily rely on the correctness of intermediate steps. As question complexity increases, spurious correlations become more prevalent, causing the reasoning process to deviate from the original intent and potentially leading to failure.
In contrast, EnoTab proactively filters spurious correlations before reasoning begins, effectively mitigating such issues and improving overall robustness. Notably, while prior work uses decomposition to construct reasoning paths, EnoTab leverages it as a supporting technique to enable fine-grained assessment of semantic units in complex questions.

\begin{table}[t]
\centering
\resizebox{0.5\textwidth}{!}{
\begin{tabular}{l|cc|c}
\toprule
\multirow{2}{*}{\textbf{Dataset}}  & \multicolumn{2}{c|}{\textbf{\# Tokens per Table}} &\multirow{2}{*}{\textbf{Comp. (\%)}} \\
\cmidrule(lr){2-3} 
& \textbf{Entire Table} & \textbf{Pruned Table} & \\
\midrule
TabFact & 343  & 216  & 37.0\%  \\
WikiTQ & 627  & 294   & 53.1\%  \\
STQA-L  & 8,967 & 2,176 & 75.7\%  \\
STQA-N & 26,742  & 2,893 & 89.2\%  \\
\bottomrule
\end{tabular}
}
\caption{Token counts and compression rates before and after pruning across four datasets.}
\label{tab:compression}
\end{table}

\paragraph{Table Pruning}
Another line of research focuses on pruning tables by generating SQL or Python programs to filter out irrelevant data, producing smaller subtables that reduce reasoning complexity \citep{zhang2023reactable,nahid2024tabsqlify,abhyankar2025h,nahid2024normtab,mao2024potable}. However, these methods often operate as black boxes, lacking transparency and making it difficult to detect and correct errors in time, which can result in the loss of critical answer-related data. In contrast, EnoTab ensures that each pruning step is observable and verifiable, allowing timely detection and correction of erroneous states.

\section{Conclusion}
In this paper, we identify the bottlenecks of TableQA in handling complex questions and noisy large-scale tables, which stem from limited relevance filtering and pruning. To tackle these challenges, we present EnoTab, a dual denoising framework that decomposes questions into semantic units, grounds them in tables, and builds explicit reasoning paths with rollback to preserve answer-related information. Extensive experiments across multiple benchmarks demonstrate that EnoTab achieves substantial gains on complex and large-scale TableQA tasks, highlighting the effectiveness and generality of our approach.

\section*{Limitations}
EnoTab is currently evaluated in the context of single-table question answering. Although the method is capable of handling multiple tables, its performance in multi-table settings remains unclear. Future work will explore and assess its effectiveness in such scenarios. Additionally, the validation method struggles with certain specialized table formats, such as "\textit{1-1}", where the numbers represent the count of wins and losses. We aim to address this limitation in future work.

\section*{Ethics Statement}
All datasets used in this study are publicly available through peer-reviewed publications cited in the references. Our framework incorporates GPT-4o and GPT-4o-mini, which may inherit ethical concerns associated with large language models, such as the potential to generate inaccurate or harmful content. We encourage users to critically evaluate the outputs produced by EnoTab before downstream use. Additionally, our method leverages open-source models such as Qwen-1.5 and Llama-2. We adhere to their usage policies and license agreements, and we acknowledge their significant contribution to this research.

\section*{Acknowledgement}

This work is supported by the National Key R\&D
Program of China, No. 2024YDLN0004 and the National Natural Science Foundation of China (U23A20275).

\bibliography{anthology,custom}
\bibliographystyle{acl_natbib}

\appendix

\section{Additional Implementation Details of EnoTab}
\label{app:impl}
This section provides additional implementation details of EnoTab to supplement the high-level overview in the main text. We elaborate on the design of key components, including evidence generation, evidence evaluation, Evidence Tree construction, and rollback, along with the associated prompting strategies and parameter configurations. These details aim to clarify the complete pipeline and support both transparency and reproducibility.

\subsection{Two-Stage Retrieval Process}

During the evidence generation phase, directly using the entire table to produce evidence is computationally inefficient and prone to errors, particularly when the table contains thousands of rows. Selecting a small set of representative rows that are most relevant to the question is typically sufficient for generating high-quality evidence. To achieve this, we adopt a two-stage retrieval strategy that balances efficiency and accuracy. The process consists of keyword-based filtering followed by embedding-based re-ranking.

\paragraph{Stage 1: Keyword-based Filtering.}  
Given a question $Q$, we first extract a set of task-relevant keywords $\{k_1, k_2, \ldots, k_m\}$ using a lightweight language model. We then apply LSH to retrieve candidate rows $R_c$ from the table $T$ that exhibit substantial token overlap with the extracted keywords. This step significantly reduces the candidate space from thousands of rows to a much smaller subset.

\paragraph{Stage 2: Embedding-based Re-ranking.}  
For each candidate row $r \in R_c$, we compute its semantic similarity with the question $Q$ using a pre-trained embedding model $f(\cdot)$. To enhance robustness, we also incorporate a lexical similarity score based on the edit distance between the row tokens and the extracted keywords. The final ranking score for row $r$ is defined as:
\[
\text{Score}(r, Q) = \lambda \cdot S_{\text{sem}}(r,Q) + (1 - \lambda) \cdot S_{\text{lex}}(r,Q),
\]
where $S_{\text{sem}}$ denotes semantic similarity measured via cosine similarity of embeddings, and $S_{\text{lex}}$ denotes lexical similarity based on edit distance.

In practice, we set $k = 10$, $\lambda = 0.7$, and $C = \min(256, \lceil 0.1N \rceil)$, where $N$ is the number of rows in the table. We use GPT-4o-mini for keyword extraction and \texttt{bge-large-en-v1.5} as the default embedding encoder. This two-stage retrieval strategy significantly improves the efficiency of evidence generation.

\subsection{Consistency Assessment of Evidence}
\label{appendix:A.2}
The consistency assessment process of evidence is presented in Algorithm~\ref{alg:evidence_consistency}.
Following \citep{lin2025explore}, we set the number of rounds to $n=5$ and the threshold to $\alpha=0.8$. For semantic discrimination, we adopt Llama-2-7b-chat-hf as the default discriminator $M_d$, which strikes a balance between accuracy and efficiency. 

\begin{algorithm}[ht!]
\caption{Evidence Consistency Assessment}
\label{alg:evidence_consistency}
\KwIn{$n$ rounds of evidence sets $\mathcal{E} = \{E_1, E_2, ..., E_n\}$, threshold $\alpha$, semantic discriminator $M_d$}
\KwOut{Candidate evidence set $E_c$}
\BlankLine
Initialize empty multimap $\mathcal{G}$ for grouping evidence by (area, action)\; 
$E_c \gets \emptyset$\; 
\ForEach{evidence set $E_i \in \mathcal{E}$}{
    \ForEach{evidence $e \in E_i$}{
        Extract $(\text{area}_e, \text{condition}_e, \text{action}_e)$\;
        Append $e$ to group $\mathcal{G}[(\text{area}_e, \text{action}_e)]$\;
    }
}
\ForEach{group $G_j$ in $\mathcal{G}$}{
    \ForEach{evidence $e_i \in G_j$}{
        $c \gets 0$\;
        \ForEach{evidence $e_k \in G_j$, $e_k \neq e_i$}{
            \If{$M_d(\text{condition}_{e_i}, \text{condition}_{e_k}) = \text{True}$}{
                $c \gets c + 1$\;
            }
        }
        Compute consistency score $S(e_i) \gets \frac{c}{|G_j| - 1}$\; 
        \If{$S(e_i) \geq \alpha$}{
            Add $e_i$ to $E_c$\;
        }
    }
}
\KwRet{$E_c$}\;
\end{algorithm}

\subsection{Usability Assessment of Evidence}
\begin{figure}[ht]
    \centering
    \includegraphics[width=1.0\linewidth]{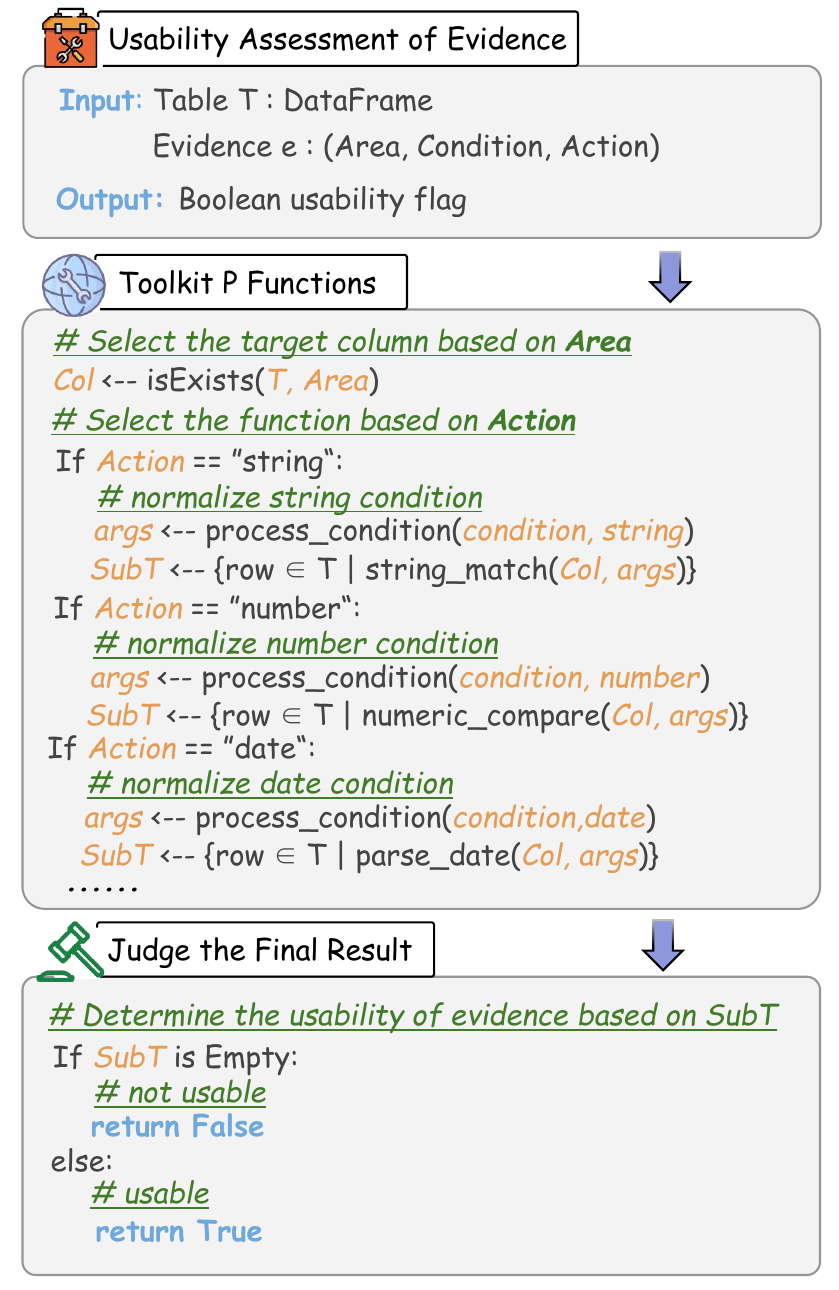}
    \caption{Usability Assessment of Evidence.}
    \label{fig:algorithm2}
\end{figure}

In this section, we describe how to verify the \emph{usability} of each evidence using the toolkit $P$ (as shown in Figure~\ref{fig:algorithm2}). 
Given an evidence $e = (\text{area}, \text{condition}, \text{action})$ and a table $T$, toolkit $P$ checks whether $e$ can be grounded in $T$. 
The process is as follows: (1) select the target column according to \texttt{area}; (2) normalize the \texttt{condition} into the canonical form required by the selected action; (3) invoke the API corresponding to \texttt{action}, such as string matching, numeric comparison, or date evaluation; (4) return whether the resulting sub-table is empty. 
If the result is non-empty, the evidence is regarded as \emph{usable}; otherwise, it is discarded. 

\subsection{Detailed Description of the Evidence Tree}
\begin{figure}[ht]
    \centering
    \includegraphics[width=1.0\linewidth]{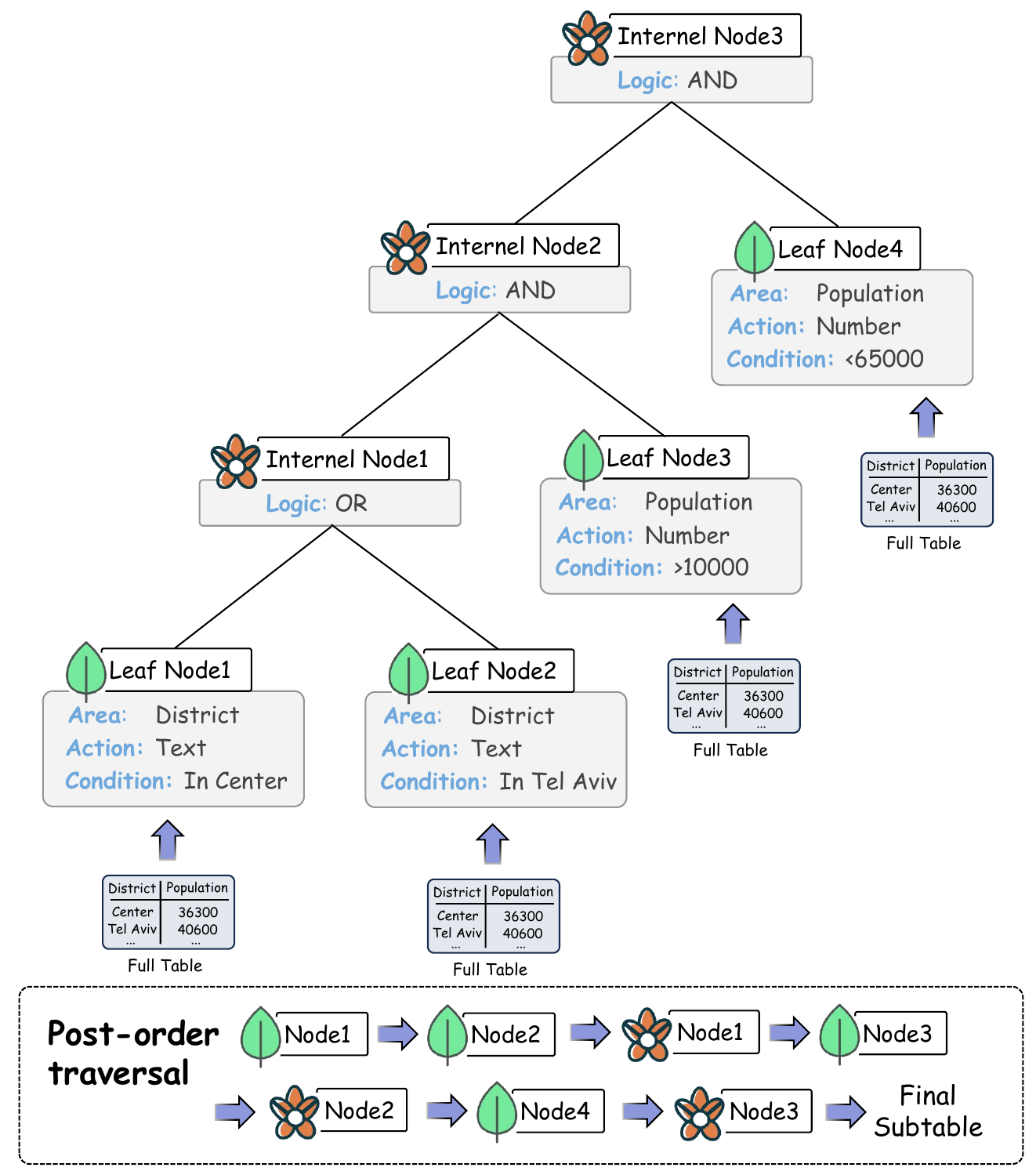}
    \caption{An illustrative Evidence Tree with four leaf nodes and three internal nodes.}
    \label{fig:tree1}
\end{figure}

In this section, we provide a more intuitive explanation of how the Evidence Tree operates. As illustrated in Figure~\ref{fig:tree1}, the Evidence Tree is structured as a binary tree consisting of four leaf nodes and three internal nodes. The entire tree mirrors the logical structure implicitly contained in the question and is executed following a post-order traversal strategy.

\begin{figure}[ht]
    \centering
    \includegraphics[width=0.86\linewidth]{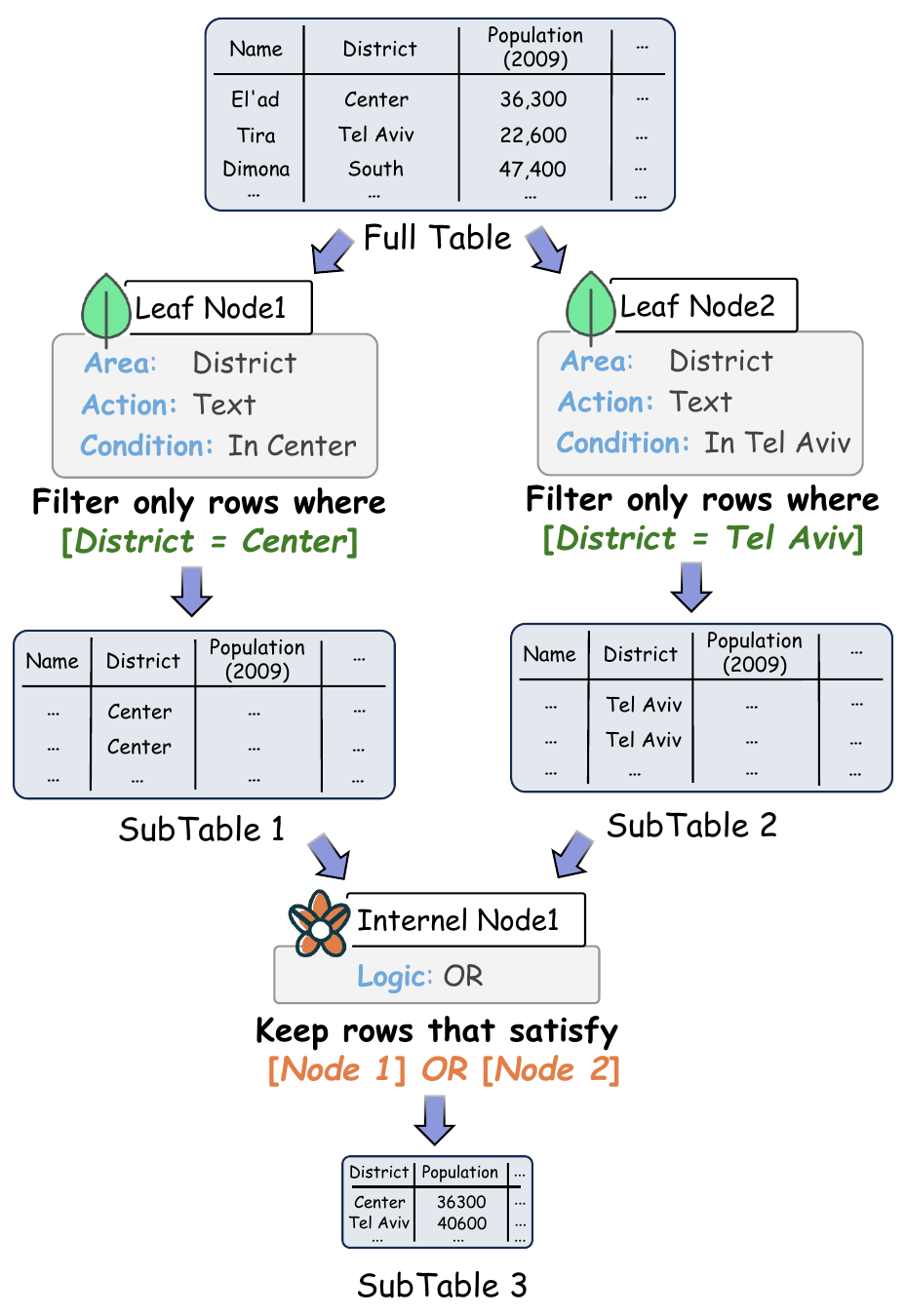}
    \caption{Table transformation process at leaf and internal nodes of the Evidence Tree.}
    \label{fig:tree2}
\end{figure}

Figure~\ref{fig:tree2} further illustrates the functional role of leaf nodes and internal nodes. Each leaf node corresponds to a minimal evidence unit and applies a fine-grained filtering operation on the original table. The internal nodes act as logical operators (e.g., AND, OR) that merge the filtered subtables produced by their children. Through this layered design, the pruning process becomes both stepwise and interpretable: every pruning decision is transparent and can be independently inspected.

Importantly, all leaf-level filtering is performed before any internal merging, which avoids prematurely combining incomplete subtables. At each step, the intermediate subtable remains observable, enabling timely detection of abnormal states (e.g., empty tables) and facilitating recovery strategies such as rollback. This property distinguishes the Evidence Tree from black-box pruning approaches and ensures that the pruning trajectory remains auditable and robust.

\subsection{Rollback Mechanism: Analysis of And2Or}
The rollback mechanism is designed to prevent the loss of essential information during table pruning. In particular, the And2Or operation acts as a fallback strategy when an internal node with an AND operator produces an empty subtable. This situation indicates that the system has reached the limits of its discriminative capability, being unable to determine which records are relevant. To avoid discarding all potentially useful data, the AND operator is replaced with an OR operator, resulting in a superset that may include redundant or irrelevant rows. While this relaxation may reduce precision, it preserves answer-critical content and ensures that downstream reasoning has sufficient evidence to proceed. In the context of TableQA, recall is generally more important than precision, as irrelevant rows can be filtered later, whereas missing evidence cannot be recovered. In practice, the And2Or operation helps maintain non-empty intermediate results, allowing later verification modules to assess completeness and trigger corrections when necessary. This mechanism contributes to greater robustness compared to approaches that discard all content under overly strict conditions. In future work, we aim to enhance the model's discriminative ability at conjunction nodes, reducing reliance on And2Or while preserving its effectiveness as a safeguard.

\subsection{Details of Dataset Construction}
\label{apx:data}

We construct two evaluation sets, \textsc{STQA-L} and \textsc{STQA-N}, from source tables through a semi-automatic QA generation pipeline with manual verification. For each source table, we randomly sample a region, which can be a single cell, a row, or a sub-table consisting of multiple rows and/or columns. The sampled region is not directly used as the final answer. Instead, we prompt an LLM to generate an answer conditioned on the sampled region and then construct a corresponding question. All generated QA pairs are manually reviewed and verified for correctness. Problematic cases identified during review are either corrected or discarded. More than 80\% of QA pairs are accepted without edits, while fewer than 20\% require manual refinement. We consider three sampling granularities in this process: cell-based sampling selects a single cell, row-based sampling selects an entire row, and subtable-based sampling selects a multi-row, multi-column sub-table.

For \textsc{STQA-N}, we first expand tables shorter than 4,096 tokens by prompting an LLM to generate additional rows and columns, and then construct QA pairs using the same procedure as in \textsc{STQA-L}. After identifying the gold answer cells and columns, we perturb 15\% of the non-answer cells with type-specific noise. For string cells, including dates and booleans, we replace the original value with one of the top five semantically closest WordNet alternatives. For numeric cells, we replace the original value with a value outside the gold range while keeping answer correctness unchanged. All QA pairs are then manually verified to ensure that the injected noise does not compromise validity. This process creates realistic distractors and enables evaluation of noise sensitivity.

During QA construction, we observe two major error types. The first is \textit{Answer Error}, where the generated question is correct but the answer is inaccurate; such QA pairs are retained and manually corrected based on the table contents. The second is \textit{Invalid Question}, where the question is ill-posed, ambiguous, or inconsistent with the table; such QA pairs are discarded entirely. Table~\ref{tab:qa_error_stats} summarizes the statistics of QA construction outcomes, and Table~\ref{tab:dataset} reports the final dataset statistics.

\begin{table}[t]
\centering
\small
\setlength{\tabcolsep}{5pt}
\begin{tabular}{lccc}
\toprule
\textbf{Dataset} & \textbf{Correct} & \textbf{Answer Error} & \textbf{Invalid Question} \\
\midrule
\textsc{STQA-L} & 81.2\% & 12.7\% & 6.1\% \\
\textsc{STQA-N} & 80.7\% & 16.4\% & 2.9\% \\
\bottomrule
\end{tabular}
\caption{Statistics of QA construction outcomes.}
\label{tab:qa_error_stats}
\end{table}

\begin{table}[t]
\centering
\small
\setlength{\tabcolsep}{6pt}
\begin{tabular}{lccc}
\toprule
\multirow{2}{*}{\textbf{Dataset}} & \multirow{2}{*}{\textbf{\# QA Pairs}} & \textbf{\# Tokens} & \textbf{\# Tokens} \\
 &  & \textbf{per Table} & \textbf{per Answer} \\
\midrule
\textsc{STQA-L} & 1,074 & 9,786 & 3.9 \\
\textsc{STQA-N} & 617 & 28,652 & 3.1 \\
\bottomrule
\end{tabular}
\caption{Statistics of QA pairs and token counts per table and answer.}
\label{tab:dataset}
\end{table}

\section{Additional Experiments}
\subsection{Hyper-parameter Settings}
\begin{table}[ht]
\centering
\resizebox{0.48\textwidth}{!}{
\begin{tabular}{ll}
\toprule
\textbf{Description} & \textbf{Value / Model} \\
\midrule
Top-$k$ representative rows & $k = 10$ \\
Candidate pool size & $C = \min(256, \lceil 0.1N \rceil)$ \\
Ranking weight & $\lambda = 0.7$ \\
Keyword extraction model & GPT-4o-mini \\
Embedding encoder & \texttt{bge-large-en-v1.5} \\
Consistency rounds & $n = 5$ \\
Consistency threshold & $\alpha = 0.8$ \\
Semantic discriminator & Llama-2-7b-chat-hf \\
\bottomrule
\end{tabular}
}
\caption{Summary of hyper-parameter settings for EnoTab.}
\label{tab:hyperparams}
\end{table}

Table~\ref{tab:hyperparams} summarizes the models and parameter settings used in each module of EnoTab. To ensure the stability of our results, we also repeated experiments with different random seeds. For open-source models (e.g., LLaMA-2, Qwen-1.5), we report the average accuracy over 3 runs. For closed-source APIs (e.g., GPT-4o, GPT-4o-mini), which show minimal stochastic variation, we conducted 2 runs to confirm stability. Across all datasets, the results were highly stable, with the standard deviation consistently within 1\% absolute accuracy. This demonstrates that the reported improvements of EnoTab are robust and not dependent on random factors.

\begin{table}[t]
  \centering  
    \begin{tabular}{l|cccc}
    \toprule
    \multirow{2}[4]{*}{\textbf{Method}} & \multicolumn{4}{c}{\textbf{FeTaQA}} \\
    \cmidrule{2-5}      & \textbf{BLEU} & \textbf{R-1} & \textbf{R-2} & \textbf{R-L} \\
    \midrule
    T5-small & - & 55 & 33 & 47\\
    T5-base & - & 61 & 39 & 51\\
    T5-large & - & 63 & 41 & 53\\
    End-to-End QA & 28.37 & 63  & 41  & 53 \\
    Dater & 29.47 & 63 & 41 & 53 \\
    Chain-of-Table & 32.61 & 66 & 44 & 56\\
    \midrule
    \textbf{EnoTab(ours)} & 
    \textbf{30.46} & 
    \textbf{67}  & 
    \textbf{45}  & 
    \textbf{57} \\
    \bottomrule
\end{tabular}%
  \caption{Performance comparison between EnoTab and previous work on the FeTaQA dataset (evaluated using GPT-3.5-turbo). BLEU and ROUGE scores are reported: R-1, R-2, and R-L denote ROUGE-1, ROUGE-2, and ROUGE-L respectively.}
  \label{tab:results-fetaqa}
\end{table}



\subsection{Error Analysis}
\begin{figure}[ht]
    \centering
    \includegraphics[width=1.0\linewidth]{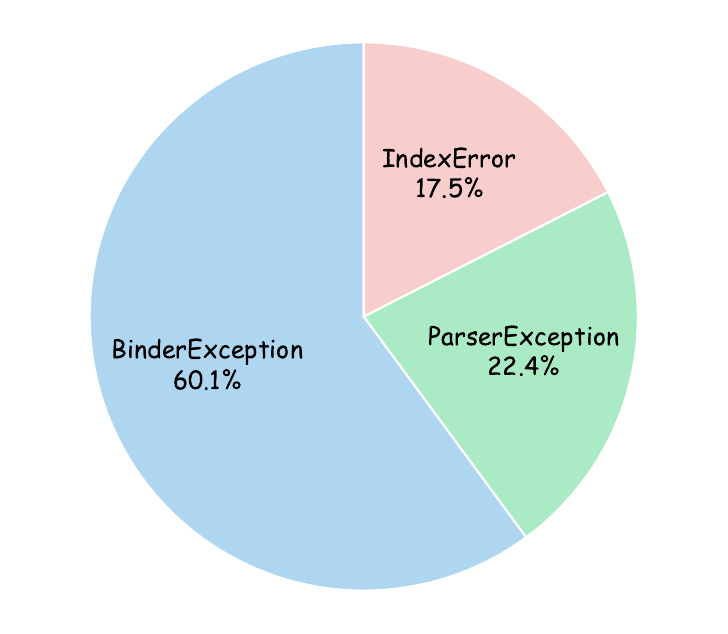}
    \caption{Statistics for the SQL execution errors.}
    \label{fig:error}
\end{figure}

Figure~\ref{fig:error} presents our error analysis of the traditional table pruning method Text2SQL on the WikiTQ dataset. The errors are categorized into three types: BinderException, which indicates semantic errors such as referencing non-existent columns, tables, or aliases; ParserException, which refers to SQL syntax errors that prevent parsing; and IndexError, which occurs when execution attempts to access out-of-range rows (e.g., accessing row 10 in a 5-row table). Among these, BinderException is the most frequent. This type of error often arises when any condition in the SQL query is invalid, rendering the entire query unusable and causing pruning to fail (see case in Figure~\ref{fig:case1}). ParserException is the second most common and typically results from conditions or table content that exceed the expressive capacity of SQL (see case in Figure~\ref{fig:case2}).

Our proposed method, EnoTab, effectively addresses these issues. Since each evidence unit is treated as an independent and minimal semantic element, EnoTab decouples multiple SQL conditions into discrete pieces of evidence. Each piece is individually verified and executed, allowing invalid evidence to be filtered out and significantly reducing the occurrence of BinderException. Moreover, EnoTab is able to recognize when certain evidence exceeds its table reasoning capability, thereby avoiding ParserException by design.

While EnoTab is robust in handling large-scale tables, its performance is still limited when dealing with structurally complex tables. For example, some cells contain compound values such as “1–1” (indicating 1 win and 1 loss) or “251–32=189” (where 189 is the value of interest). Successfully pruning such tables requires accurately extracting the relevant data from compound entries. Although equipping the model with this ability could further improve pruning effectiveness, it also introduces a higher risk of losing critical answer-related data. This conflicts with the core principle of EnoTab: preserving target data while pruning as aggressively as possible.

\begin{figure}[t]
    \centering
    \includegraphics[width=0.96\linewidth]{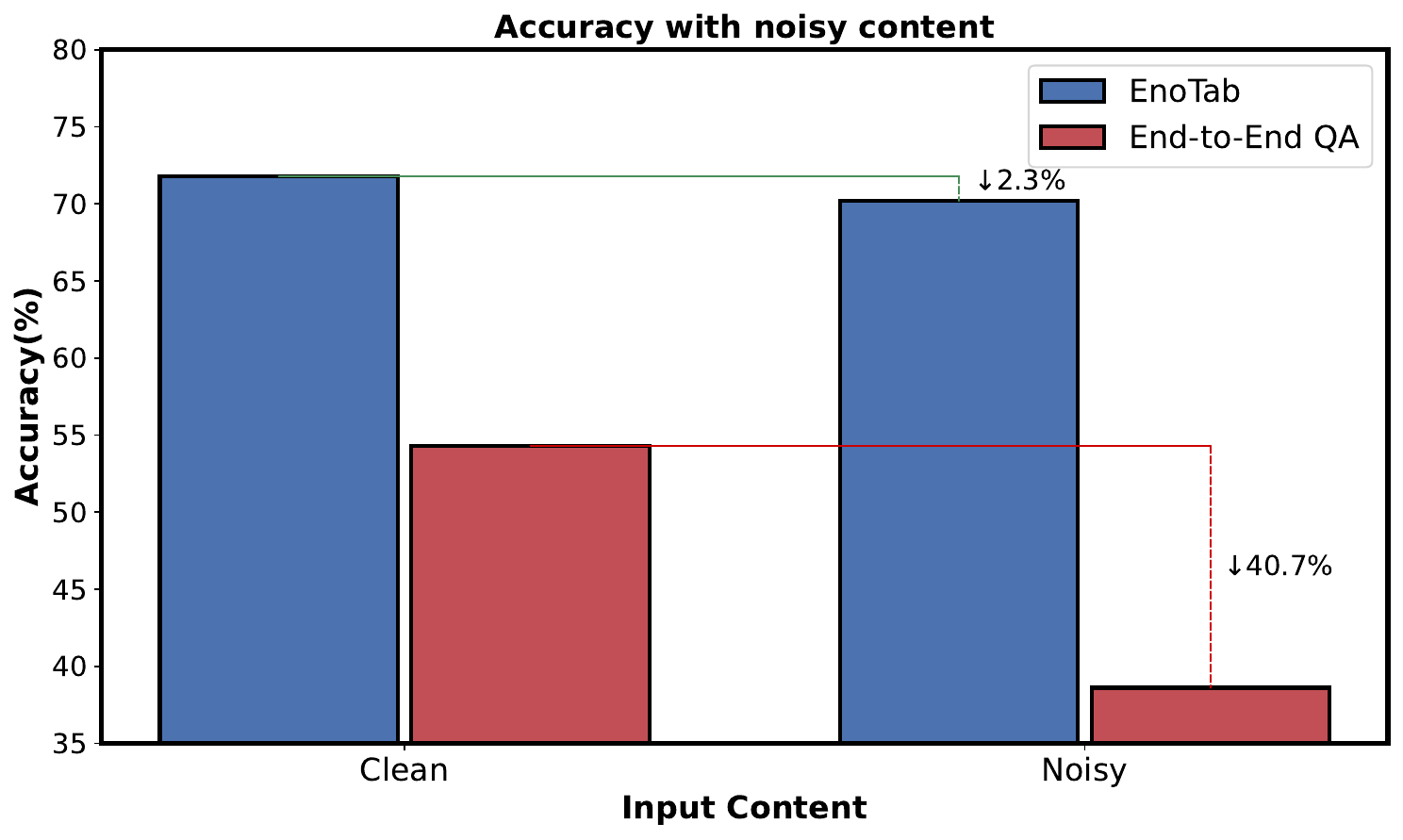}
    \caption{Execution accuracy on WikiTQ with noisy content in tables.}
    \label{fig:noisy}
\end{figure}

\subsection{Noisy Content}
To further evaluate EnoTab’s robustness to noisy data, we follow the approach proposed in Binder\citep{Binder} and construct a noisy version of the WikiTQ development set by injecting distractive content to simulate misleading information in real-world scenarios. We evaluate both EnoTab and a standard End-to-End QA method under two settings: “clean” and “noisy”.
As shown in Figure~\ref{fig:noisy}, the performance of End-to-End QA drops significantly when noise is introduced, indicating its sensitivity to misleading content. In contrast, EnoTab maintains stable performance even under noisy conditions, demonstrating stronger robustness. This advantage primarily stems from its efficient table pruning capability, which enables the model to effectively identify and filter out irrelevant information, thereby sustaining high reasoning accuracy even in the presence of noise.

\subsection{Experiments of EnoTab on FetaQA}
Table~\ref{tab:results-fetaqa} shows that EnoTab achieves strong performance on the FeTaQA\citep{nan2022fetaqa} dataset for free-form question answering, consistently outperforming existing state-of-the-art methods. In particular, EnoTab surpasses all baselines on ROUGE-1/2/L\citep{lin2004rouge}, which measure lexical overlap (unigram and bigram) and sequence-level similarity based on the longest common subsequence. We attribute this improvement to EnoTab's ability to better capture key information from the table and maintain structural alignment in the generated answers through fine-grained evidence selection and reasoning.
We also observe that EnoTab slightly underperforms Chain-of-Table\citep{wang2024chain} in terms of BLEU\citep{papineni2002bleu} score. Since BLEU places greater emphasis on n-gram precision and is sensitive to word order, we believe this is due to the fact that our model does not explicitly optimize for surface-level phrasing or sequence alignment. However, manual inspection confirms that the generated answers remain accurate and complete in content, indicating that the performance drop in BLEU does not reflect true semantic degradation.

\begin{table}[ht]
  \centering
  \footnotesize
\begin{tabular}{lcccc}
\toprule
\multirow{2}[4]{*}{\textbf{Dataset}} & \multicolumn{4}{c}{\textbf{Evidence Count}} \\
\cmidrule{2-5}  & \textbf{1} & \textbf{2} & \textbf{3} & \textbf{\textgreater 4} \\
\midrule
{WikiTQ} & 1628   & 1277 & 1108   & 296  \\
{TabFact} & 1134   & 573   & 287   & 29 \\
\bottomrule
\end{tabular}
\caption{Number of samples in the WikiTQ and TabFact dataset by evidence count.}
  \label{tab:evidence_count}%
\end{table}%

\begin{table}[ht]
  \centering
  \footnotesize
\begin{tabular}{lcccc}
\toprule
\multirow{2}[4]{*}{\textbf{Dataset}} & \multicolumn{4}{c}{\textbf{Difficulty Level}} \\
\cmidrule{2-5}  & \textbf{easy} & \textbf{middle} & \textbf{hard} & \textbf{extra} \\
\midrule
{WikiTQ} & 1572   & 1374 & 1003   & 360  \\
{TabFact} & 1347   & 463   & 171   & 42 \\
\bottomrule
\end{tabular}
\caption{Number of samples in the WikiTQ and TabFact dataset by difficulty level.}
  \label{tab:difficulty level}%
\end{table}%


\subsection{Effectiveness Analysis of Evidence Generation}
To evaluate the effectiveness of our evidence generation module, we analyze the relationship between the number of generated evidence pieces and task difficulty across two benchmarks: WikiTQ and TabFact. As shown in Tables~\ref{tab:evidence_count} and~\ref{tab:difficulty level}, the distribution of samples across different evidence counts closely mirrors the distribution across difficulty levels. This alignment suggests that tasks requiring more evidence are generally more difficult.
More notably, we observe that the number of samples with higher evidence counts increases with task difficulty. For instance, in WikiTQ, the proportion of samples requiring 4 or more evidence pieces rises as we move from 'easy' to 'extra' difficulty levels. A similar trend is evident in TabFact, albeit with fewer high-difficulty samples. This increasing evidence requirement reflects the heightened complexity of reasoning in harder tasks and demonstrates that our evidence generation module is sensitive to task difficulty.
These findings confirm that our module can effectively capture and extract more granular, informative evidence as the reasoning complexity increases. In other words, it scales appropriately with task demands and is particularly beneficial for handling more challenging questions.



\section{Prompt}
\label{apx:prompt}

\subsection{Prompt of Dataset Construction}
\begin{figure*}[ht]
    \centering
    \includegraphics[width=1.0\linewidth]{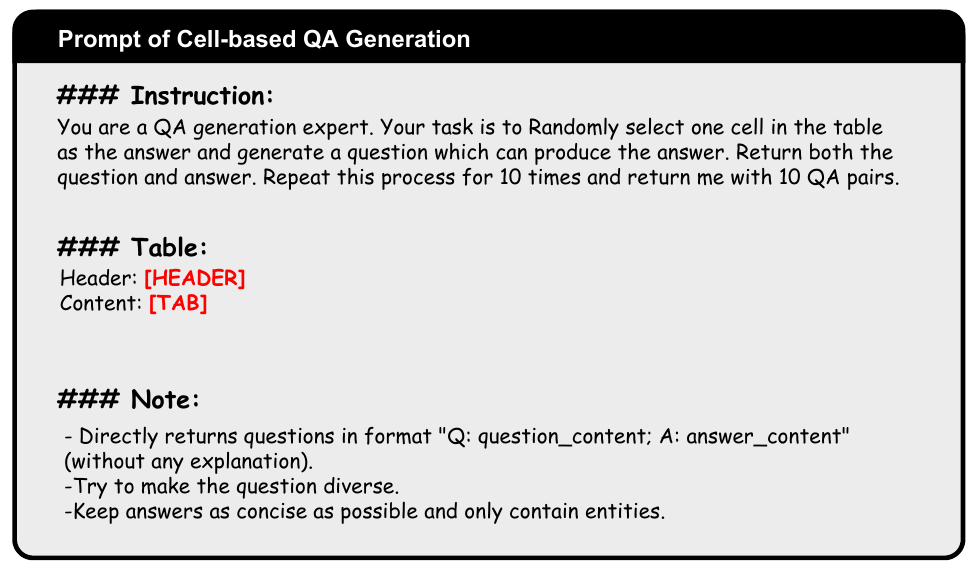}
    \caption{Prompt of Cell-based QA Generation.}
    \label{fig:data_cell}
\end{figure*}

\begin{figure*}[ht]
    \centering
    \includegraphics[width=1.0\linewidth]{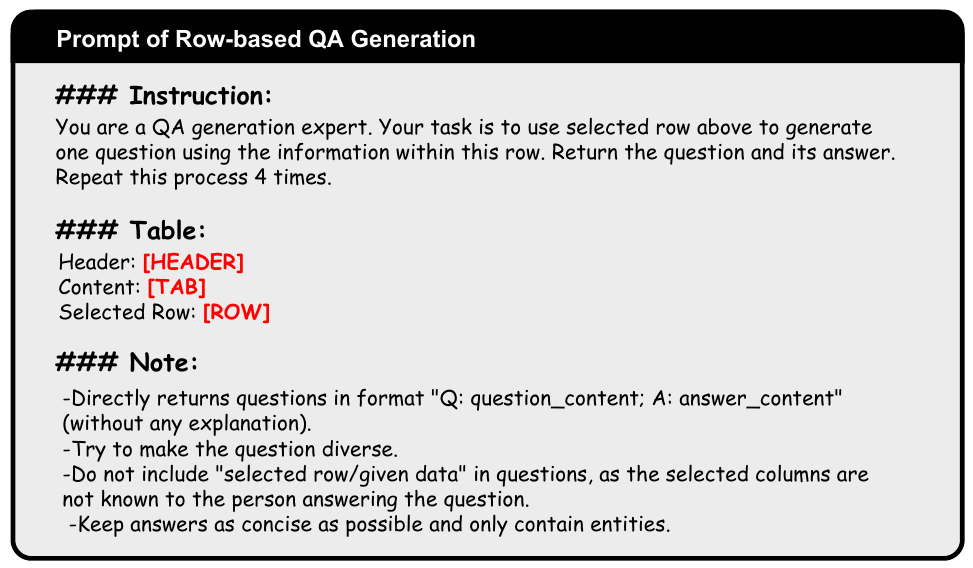}
    \caption{Prompt of Row-based QA Generation.}
    \label{fig:data_row}
\end{figure*}

\begin{figure*}[ht]
    \centering
    \includegraphics[width=1.0\linewidth]{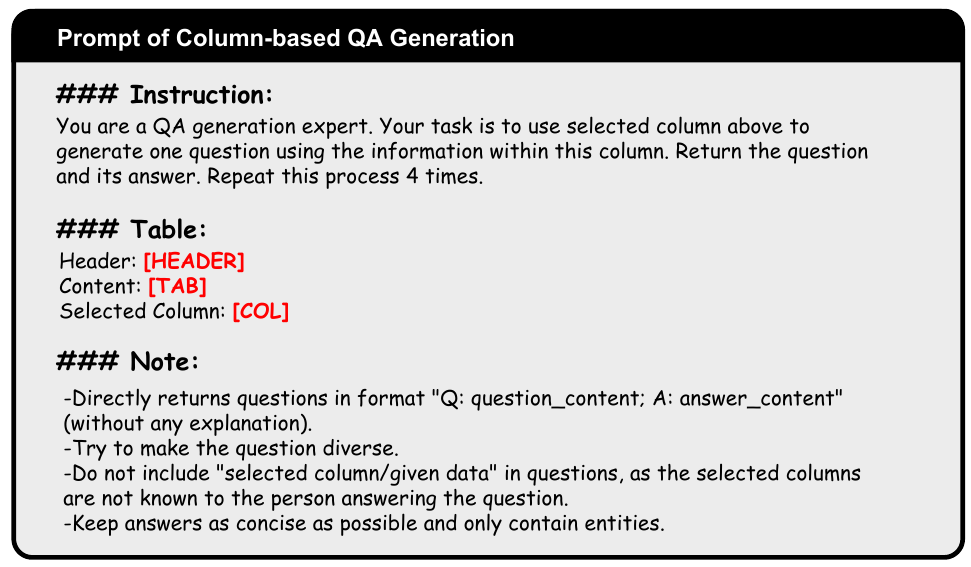}
    \caption{Prompt of Colmn-based QA Generation.}
    \label{fig:data_col}
\end{figure*}

\begin{figure*}[ht]
    \centering
    \includegraphics[width=1.0\linewidth]{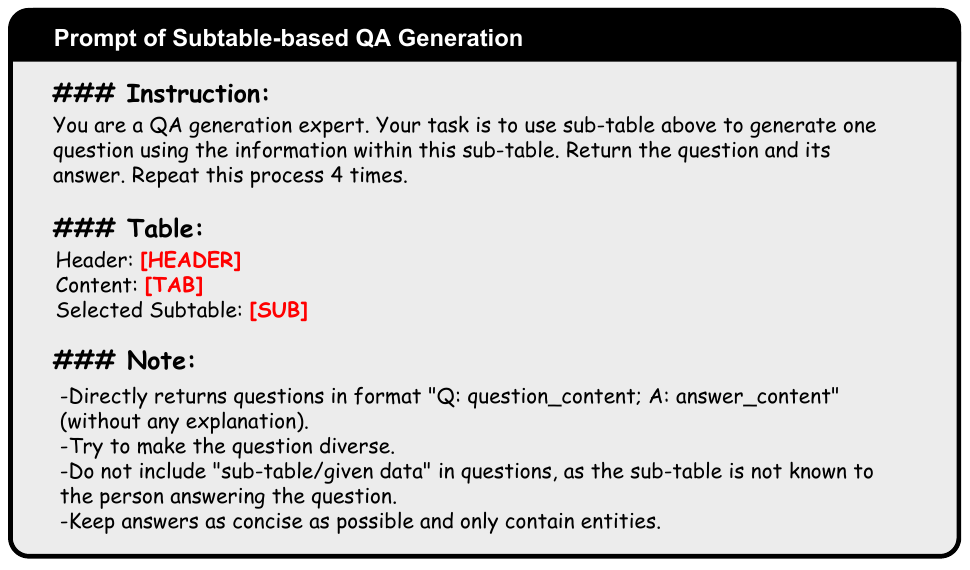}
    \caption{Prompt of Subtable-based QA Generation.}
    \label{fig:data_subtable}
\end{figure*}

\begin{figure*}[ht]
    \centering
    \includegraphics[width=1.0\linewidth]{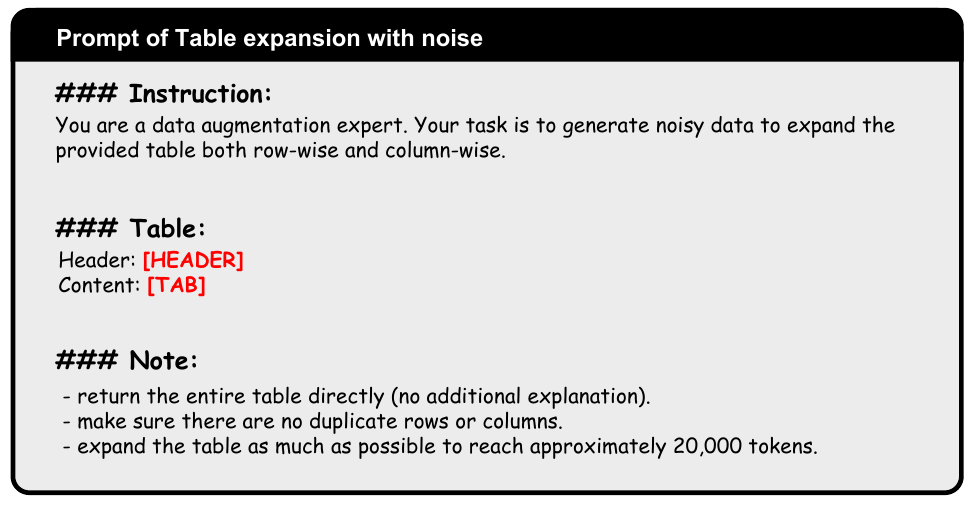}
    \caption{Prompt of Table expansion with noise.}
    \label{fig:data_expand}
\end{figure*}

The prompt used in data construction, as shown in Figure \ref{fig:data_cell}, \ref{fig:data_row}, \ref{fig:data_col}, \ref{fig:data_subtable} and \ref{fig:data_expand}.

\subsection{Prompt of EnoTab}
\begin{figure*}[ht]
    \centering
    \includegraphics[width=1.0\linewidth]{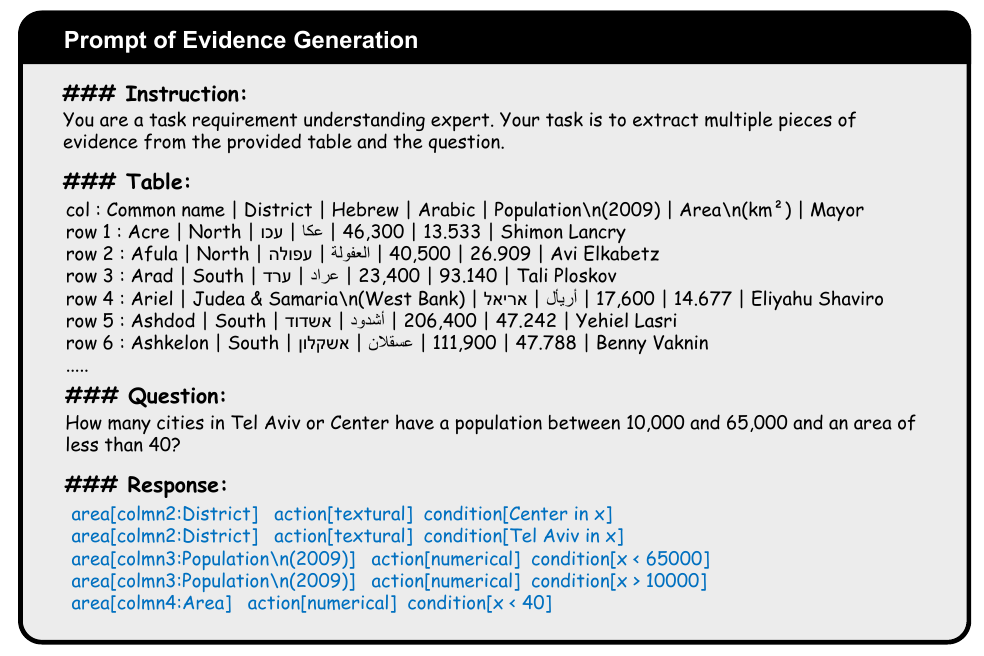}
    \caption{Prompt of Evidence Generation.}
    \label{fig:prompt_1}
\end{figure*}

\begin{figure*}[ht]
    \centering
    \includegraphics[width=1.0\linewidth]{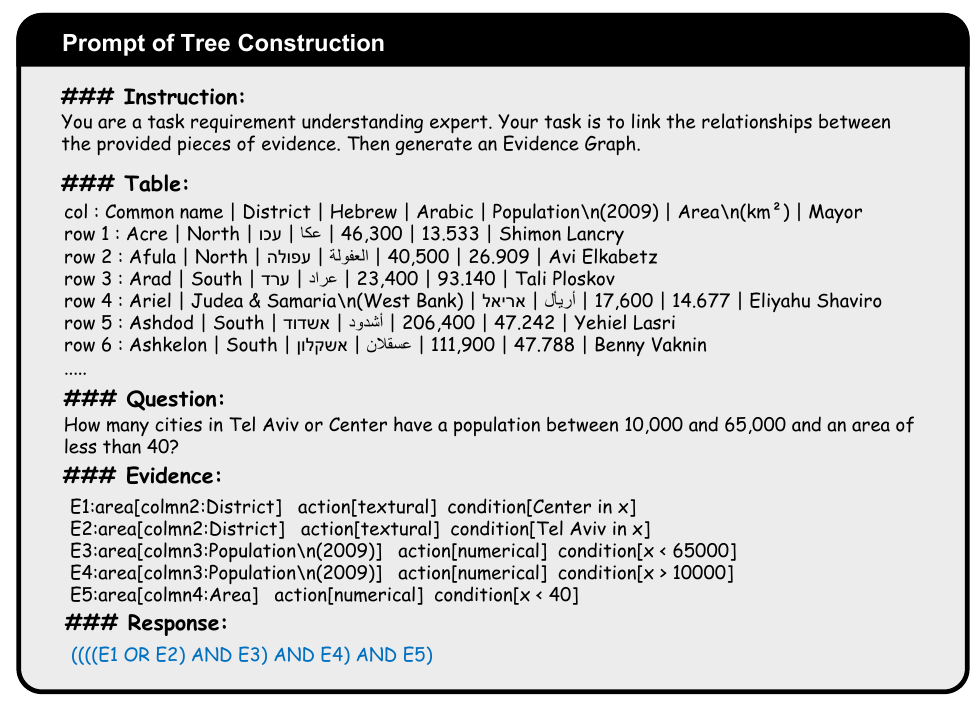}
    \caption{Prompt of Tree Construction.}
    \label{fig:prompt_2}
\end{figure*}

\begin{figure*}[ht]
    \centering
    \includegraphics[width=1.0\linewidth]{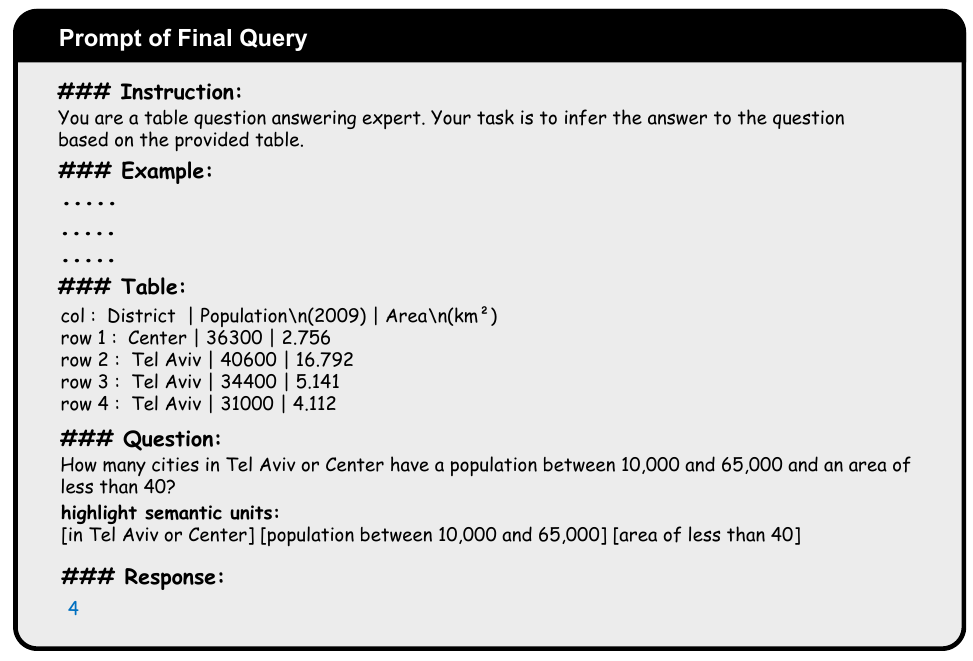}
    \caption{Prompt of Final Query.}
    \label{fig:prompt_3}
\end{figure*}

\begin{figure*}[ht]
    \centering
    \includegraphics[width=1.0\linewidth]{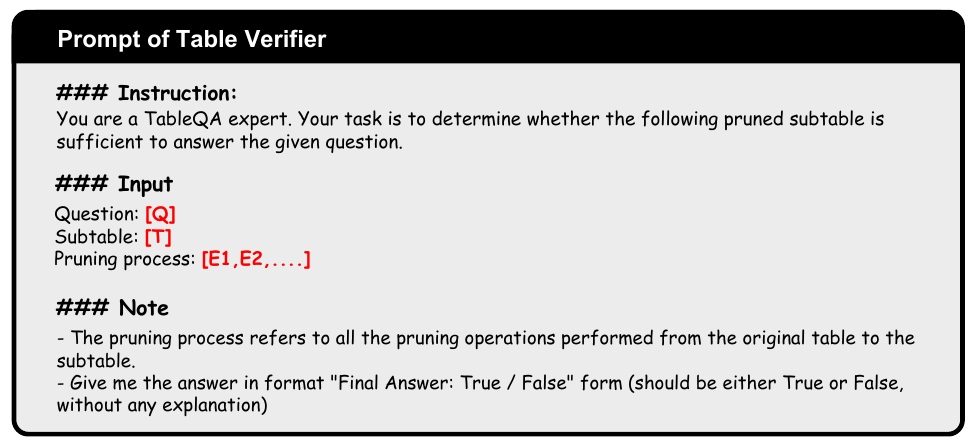}
    \caption{Prompt of Table Verifier.}
    \label{fig:prompt_4}
\end{figure*}

The prompt used in the EnoTab pipeline, as shown in Figure \ref{fig:prompt_1}, \ref{fig:prompt_2}, \ref{fig:prompt_3} and \ref{fig:prompt_4}.

\subsection{Example of Evidence}
\begin{figure*}[ht]
    \centering
    \includegraphics[width=1.0\linewidth]{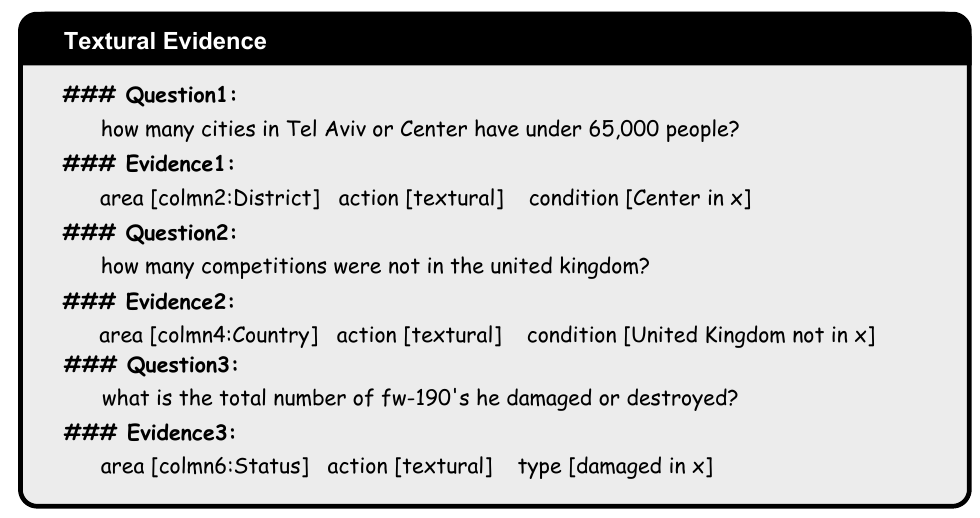}
    \caption{Example of Textural Evidence.}
    \label{fig:ex_text}
\end{figure*}

\begin{figure*}[ht]
    \centering
    \includegraphics[width=1.0\linewidth]{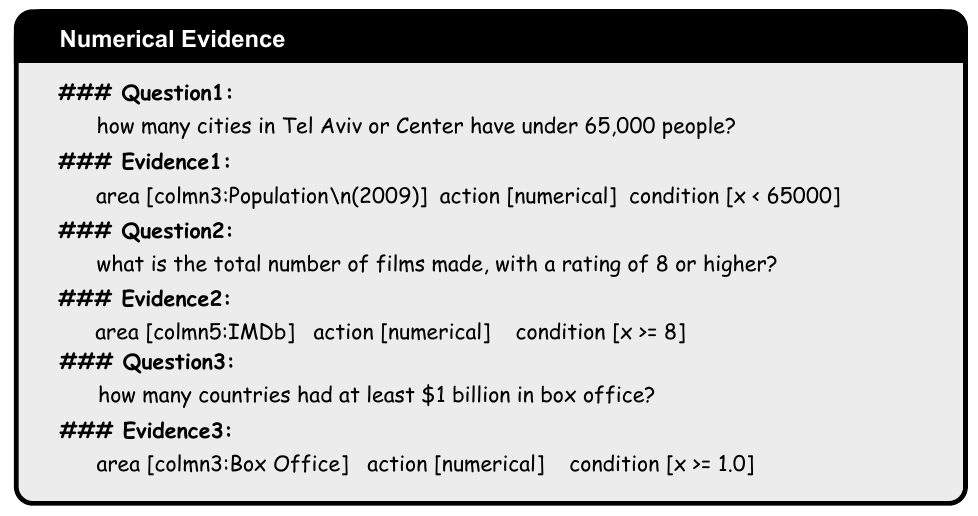}
    \caption{Example of Numberical Evidence.}
    \label{fig:ex_num}
\end{figure*}

\begin{figure*}[ht]
    \centering
    \includegraphics[width=1.0\linewidth]{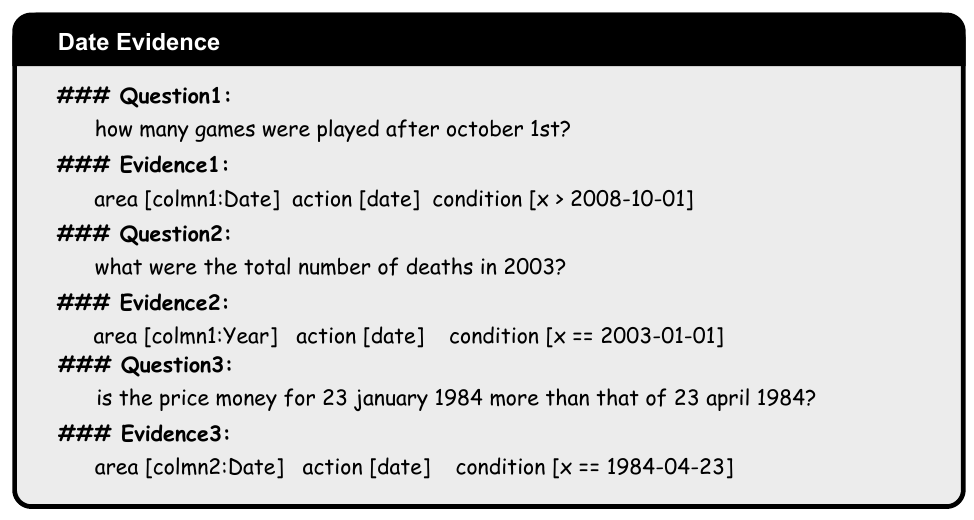}
    \caption{Example of Date Evidence.}
    \label{fig:ex_date}
\end{figure*}

\begin{figure*}[ht]
    \centering
    \includegraphics[width=1.0\linewidth]{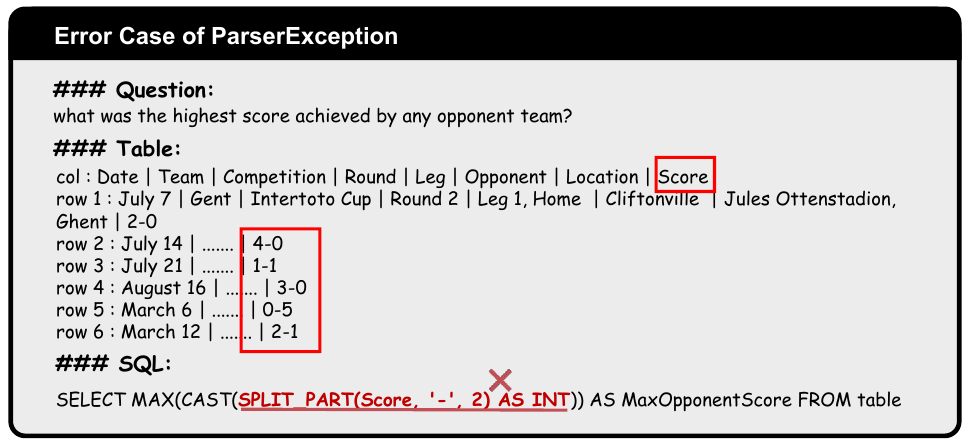}
    \caption{Error Case of ParserException.}
    \label{fig:case1}
\end{figure*}

\begin{figure*}[ht]
    \centering
    \includegraphics[width=1.0\linewidth]{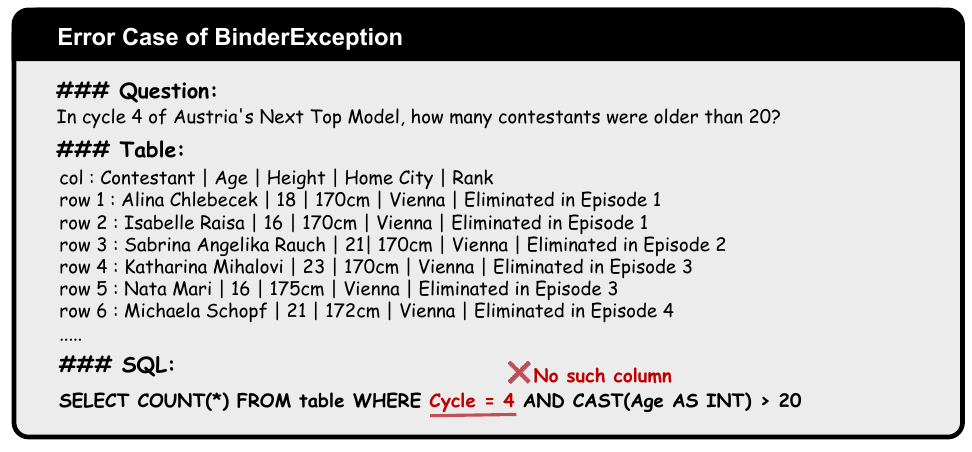}
    \caption{Error Case of BinderException.}
    \label{fig:case2}
\end{figure*}

The types of Evidence include textual, numerical and date, with corresponding examples
shown in Figures \ref{fig:ex_text}, \ref{fig:ex_num} and \ref{fig:ex_date}, respectively.

\section{Additional Related Work}
\label{apx:related_work}
\paragraph{LLM Reasoning} In recent years, the rapid progress of large language models (LLMs) has made reasoning a core capability for modern AI systems. Reasoning-based paradigms have been widely adopted across diverse applications, including vision-language modeling \citep{liu2024synthvlm,liu2025fusion,lin2026mmfinereason,lin2026scientific}, chart understanding \citep{liu2026chartverse}, mathematical problem solving \citep{wu2026step,an2025amo}, and web-based autonomous agents \citep{zhang2026expseek}. Meanwhile, emerging evidence suggests that reasoning processes are highly heterogeneous across different tokens, modules, modalities, and intermediate steps, calling for more adaptive reasoning, optimization, and inference strategies \citep{liu2025uniform,zhou2025dropping,zhou2026look}. Against this backdrop, structured reasoning tasks have attracted increasing attention, including table reasoning and closely related text-to-SQL settings, where models must jointly understand structured schemas, content, and compositional reasoning procedures over tabular evidence \citep{wu-etal-2025-ucs,wu2025mr,wu2025table}.

\paragraph{Finetune-based Table Reasoning}
In the field of table reasoning, early methods typically enhance models' understanding of tables by fine-tuning pre-trained models. The core challenge lies in enabling models to better comprehend the content and structure of tables during training. TaPas\citep{herzig2020tapas} improves understanding of tabular data by recovering masked cell information in tables. TABERT\citep{yin2020tabert} proposes the concept of content snapshots to encode the most relevant table content subsets based on the input utterance. TURL\citep{deng2022turl} focuses on table relationship understanding by introducing table context information and modeling cell semantics, significantly enhancing table semantic understanding and reasoning. TAPEX\citep{liu2021tapex} leverages a BART\citep{lewis2019bart} model to simulate a SQL executor during pre-training, equipping TAPEX with stronger table reasoning capabilities. PASTA\citep{gu2022pasta} introduces an operation-aware fact verification approach, pre-training the language model to learn common table-based operations and solve sentence-table cloze tasks synthesized from WikiTables\citep{pasupat2015compositional}, further improving reasoning capabilities. Additionally, \citep{sui2024table} addresses large-scale tables by defining a series of constraints to control table size while minimizing the loss of key information.
\paragraph{Pruning and Planning Table Reasoning} Some recent studies have attempted to jointly address table pruning and reasoning planning by integrating evidence selection with structured inference control. These approaches aim to reduce reasoning complexity by eliminating irrelevant data while also guiding the model through a well-defined reasoning path. For example, some frameworks alternate between executing partial programs and making intermediate decisions, allowing the reasoning process to adapt dynamically to the evolving context \citep{khoja2025weaver}. Other methods introduce symbolic planners or controller modules that determine the order in which subtasks—such as filtering, aggregation, or comparison—should be executed \citep{zhang2023reactable,mao2024potable}. While effective in structured reasoning, such methods often operate as black boxes, making it difficult to trace or correct errors in pruning or planning. In contrast, our method explicitly separates pruning and planning stages, and further introduces verifiable checkpoints at each step to enhance transparency and robustness.


\end{document}